\documentclass[12pt,aps,pre]{revtex4-2}

\usepackage[margin=1in]{geometry}
\usepackage{graphicx}
\usepackage{amssymb}
\usepackage{amsmath}
\usepackage{mathrsfs}
\usepackage{placeins}
\usepackage{comment}
\usepackage{natbib}
\usepackage{caption}
\usepackage{subcaption}
\usepackage{wrapfig}
\usepackage{float}
\usepackage[linesnumbered,algoruled,boxed,lined]{algorithm2e}
\bibpunct{(}{)}{,}{s}{,}{,}

\usepackage[bookmarksnumbered=true,breaklinks=true,colorlinks,citecolor=blue,linkcolor=blue]{hyperref}
\usepackage[usenames]{color}
\usepackage[normalem]{ulem} 

\newcommand{\MDG}[1]{{\color{magenta}*** #1 ***}}

\makeatletter

\usepackage{natbib}
\setcitestyle{numbers,square}

\begin{document}

\title{Symmetry reduction for deep reinforcement learning active control of chaotic  spatiotemporal dynamics}
\author{Kevin Zeng}
\author{Michael D. Graham}
\email{Email: mdgraham@wisc.edu}
\affiliation{Department of Chemical and Biological Engineering, University of Wisconsin-Madison, Madison WI 53706, USA}
\date{\today}

\begin{abstract}

%


Deep reinforcement learning (RL) is a data-driven, model-free method capable of discovering complex control strategies for macroscopic objectives in high-dimensional systems, making its application towards flow control promising. Many systems of flow control interest possess symmetries that, when neglected, can significantly inhibit the learning and performance of a naive deep RL approach.
Using a test-bed consisting of the Kuramoto-Sivashinsky Equation (KSE), equally spaced actuators, and a goal of minimizing dissipation and power cost, we demonstrate that by moving the deep RL problem to a symmetry-reduced space, we can alleviate limitations inherent in the naive application of deep RL.
We demonstrate that symmetry-reduced deep RL yields improved data efficiency as well as improved control policy efficacy compared to policies found by naive deep RL. 
Interestingly, the policy learned by the the symmetry aware control agent drives the system toward an equilibrium state of the forced KSE that is connected by continuation to an equilibrium of the unforced KSE, despite having been given no explicit information regarding its existence. I.e., to achieve its goal, the RL algorithm discovers and stabilizes an equilibrium state of the system.
 Finally, we demonstrate that the symmetry-reduced control policy is robust to observation and actuation signal noise, as well as to system parameters it has not observed before.

\end{abstract}

\maketitle


\section{Introduction} \label{Introduction}

The recent explosive growth in machine learning research has led to a large set of data-driven algorithms that map inputs to outputs by learning patterns and building inferences from the data without the need to hardcode explicit instructions. A subset of these methods are called semi-supervised learning algorithms, which learn under partial supervision through feedback from the environment. This subset is dominated by deep reinforcement learning (RL) algorithms, which, with the aid of neural networks, are particularly well-suited for tackling complex control problems with elusive optimal policies. In the past few years, deep RL has garnered the spotlight by solving complex, high-dimensional control problems and defeating the best human players in the world in games such as Go \citep{Silver2016} and DOTA II \citep{OpenAI2019}, which were once thought to be too high-dimensional to feasibly solve. 

Using a model system, the Kuramoto-Sivashinsky equation (KSE), that has chaotic dynamics as well as continuous and discrete symmetries analogous to those found in wall turbulence, the present work takes a step toward application of deep RL to control of spatiotemporally complex fluid flow problems, with the ultimate aim being to reduce energy losses in turbulent flows.

Deep RL offers a potential avenue for discovering active flow control policies for several reasons. Designing a complex active flow controller via analytical means is, in general, intractable. Given an array of sensor readings and actuators, no obvious strategy exists to analytically develop a concerted control scheme between the two sets \citep{Duriez2016}. Furthermore, the nonlinear complexity and high dimensionality of turbulent flows render real-time predictive simulations of potential possible actuations impractical. The bulk of existing active flow control policies are relatively simple, relying on oscillatory or constant actuation \citep{Schoppa1998}. These open-loop control policies are suboptimal in that they do not leverage the full potential of their action space. Model-based approaches can also face difficulties. As noted, in many settings solving the governing equations is too slow for any prediction-based method to be practical. Reduced-order models aimed to expedite the modeling process face difficulties in accurately modeling the non-stationary dynamics caused by the introduction of control, which can lead to unwanted behavior when far from the target state \citep{Bewley2001}. In fact, a well described model of the system may not always be readily available.

Finally, deep RL offers the ability to discover control strategies for macroscopic goals, such as minimizing drag over the entire system, as opposed to traditional control methods that focus on microscopic goals, such as suppressing certain vortex motions. Indeed an outstanding challenge in flow control is the identification of ideal control targets achievable in specific flow problems \citep{Bewley2001}. 

 Although current deep RL methods do not provide explicit performance guarantees, they may discover non-trivial novel control strategies that when paired with dynamical insight can serve as guides for the development of more robust novel controls. In this regard, Deep RL can also be viewed as a control strategy discovery tool in addition to a data-driven controller.

Although we are ultimately interested in controlling the drag in wall-bounded turbulent flows, the application of deep RL toward fluid dynamics and spatiotemporal chaotic systems in general still remains in its nascent stages, with a handful of advancements sprouting from various niche domains. Deep RL has been utilized to control simple chaotic dynamical systems such as the Lorenz system \citep{Gueniat2016,Verma2020}. Recently, \citep{Bucci2019} demonstrated the deep RL control of the Kuramoto-Sivashinsky equation by directing the flow from one fixed point of the system to another with a series of artificial jets. Other applications of RL involve learning the collective motion of fish \citep{Gazzola2016,Verma2018}, maximizing the range of robotic gliders \citep{Reddy2018}, and optimizing the motion of microswimmers \citep{Colabrese2017}. With regard to fluid flow control, two recent works have explored the application of RL in two-dimensional simulations of fluid flowing over bluff bodies \citep{Gueniat2016}, \citep{Rabault2019}. Using data from velocity sensors, these algorithms learned control policies to reduce skin-friction drag over the bluff body by actuating jets located on the cylinders. The flows in these studies however, were performed at laminar Reynolds numbers, where the dynamics are simple and low-dimensional. Recently, \citep{Fan2020} demonstrated the viability of deep RL to learn flow control strategies experimentally. In this work, the algorithm used data from a series of towing experiments to learn an efficient control strategy for spinning a pair cylinders downstream of a larger cylinder to reduce the drag on the entire assembly. Although promising, many of these approaches considered flow problems that exhibit low-dimensional dynamics, lack the rich system symmetries found in wall-bounded turbulent flows such as translational and reflection symmetries, and do not aim to explicitly control the time-averaged energy dissipation rate. Furthermore, these works do not focus on understanding dynamically the learned controlled strategies.


Many systems of interest for flow control have symmetries.  Deep learning approaches that respect these symmetries automatically rather than learn to approximate them from data are likely to have superior performance, and  within the deep-learning community is a growing body of work demonstrating the importance of incorporating symmetries of the learning domain into the deep neural-network (NN) models. For example, it has been observed by \citep{LencVedaldi2018} that the state-of-the-art AlexNet NN image classifier \citep{Krizhevsky2012} spontaneously learns redundant internal representations that are equivariant to flips, scalings, and rotations when trained on ImageNet data. Other works, described later in this section, have found that directly incorporating system symmetries can yield improved learning and performance results. As many flow geometries of interest possess a range of system symmetries, it is natural to incorporate these symmetries into the deep RL model, which to our knowledge, has not been demonstrated in deep RL flow control. Because the state of many flow systems can appear in a number of symmetric orientations, it is in our interest to ensure that these dynamically equivalent states are mapped to dynamically equivalent actions for dynamical and performance consistency. Fundamentally, this implies that we seek deep models that are functionally invariant/equivariant to the state-action symmetries of the system.

Typically, for a feedforward NN to obtain an invariant/equivariant functional form, it will need to implicitly learn weight-sharing constraints \citep{ShaweTaylor1993,Ravanbakhsh2017}. It is generally accepted that these invariances can be learned given sufficient training data \citep{Ravanbakhsh2017} and capacity \citep{Cybenko1989}. However, a consequence of symmetry preserving weight constraints is the substantial decrease in the number of effective free parameters \citep{SannaiTakaiCordonnier2019, Ravanbakhsh2017}. This lowers the overall network capacity, which means that for an arbitrary feedforward NN, one will need larger networks and by extension more training data and computing time to obtain desired performance and functional properties. This exacerbates an existing challenge in deep RL algorithms in that they can be expensive in terms of training data needs. For perspective, some of the most impressive successes, such as OpenAI 5, the deep RL model that defeated the best professional teams in the world in the game DOTA II, required 10 months of 770 Petaflops/s per day of training \citep{OpenAI2019}.

To ensure that the invariances/equivariances of the domain are respected in the learning task, there are primarily three solution types. The first solution type, data augmentation, is the simplest. This approach augments the training data to include additional symmetric permutations of the original training data with the goal of pressuring the model to implicitly learn equivariant representations \citep{Krizhevsky2012,Dieleman2015}. However, this method does not guarantee that the model will generalize, nor does it address the issue in a principled method.

The second solution type is to hard-code the symmetries into the network architecture itself. Some now ubiquitous architectures, such as convolutional NNs and recurrent NNs, have demonstrated success in improving performance by accounting for translational symmetries. However, for systems with complex or collections of symmetry groups, this hard-coding method requires carefully tailoring proper weight-constraints, designing non-dense connections, or incorporating new novel NN architectures \citep{SannaiTakaiCordonnier2019,KerivenPeyre2019,Ravanbakhsh2017,CohenWelling2016}.

The third solution type is accounting for system symmetries by applying symmetry transformations to the input prior to the network or to its encoding features \citep{Dieleman2016,Jaderberg2015,Ling2016}. \citep{Ling2016} demonstrated for learning models of systems with symmetry/invariance properties, such as turbulence and crystal elasticity, these models perform better when invariance properties are embedded into the training features compared to when training features were synthetically augmented with additional symmetric data.

In the present work, we opt for the third solution type, building the symmetries explicitly around the NN model, in favor of simplicity while still obtaining explicit symmetric properties.  Although there is a growing number of works seeking to ensure invariance by hard-coding novel architectures, there is yet to be a general method of applying these into arbitrary concerted network designs (e.g. how does one handle networks that feed into each other or have multiple input-types such as Actor-Critic networks?). We will demonstrate that for the control task of minimizing system dissipation and power cost for the Kuramoto-Sivashinsky equation in a parameter regime exhibiting chaotic dynamics, symmetry-reduced deep RL yields improved data efficiency, control policy efficacy, and dynamically consistent state-action mappings compared to naive deep RL. We further observe that the symmetry-reduced control policy learns to discover and target a forced equilibrium, related to a known equilibrium of the system, that exhibits low dissipation and power input cost, despite having been given no explicit information of its existence.

The remainder of this paper is divided into the following: In Section \ref{ProblemSetUp} we introduce the Kuramoto-Sivashinsky equation and the control task, as well as providing a brief review of deep RL and a discussion of the implications of the symmetries of the state-action space on the learning problem. We then conclude this section with an outline of our method of reducing the symmetry of the deep RL problem. In Section \ref{Results} we compare the performance of our symmetry-reduced deep RL to naive deep RL approaches, investigate the learned control strategy through a dynamical systems lens, and probe the robustness of the policy. Finally, we summarize our results in Section \ref{Conclusions} and provide a discussion of the extension of this work towards more realistic problems.

\section{Formulation} \label{ProblemSetUp}

\subsection{The Kuramoto-Sivashinsky Equation and Controls}

The Kuramoto Sivashinsky Equation (KSE) is given by
\begin{equation}
  u_t = -uu_{x}-u_{xx}-u_{xxxx}+f(x,t).
 \label{eq:KSE}
\end{equation}

Here $f$ is a spatio-temporal forcing term that will be used for control actuation. We consider the KSE in a domain of length $L=22$ with periodic boundary conditions as this system has been extensively studied and exhibits analogous symmetries to flow systems of interest. The uncontrolled KSE, $f=0$, exhibits rich dynamics and spatio-temporal chaos, which has made it a common toy problem and proxy system for the Navier-Stokes Equations. The equation is time evolved with a time step of $\Delta t=0.05$ using the same numerical method and code as \citep{Bucci2019} with a third-order semi-implicit Runge-Kutta scheme, which evolves the linear second and fourth order terms with an implicit scheme and the nonlinear convective and forcing terms with an explicit scheme. Spatial discretization is performed with Fourier collocation on a mesh of 64 points. We primarily consider the domain size $L=22$.

Importantly, the KSE possesses translational and reflection symmetries, which are also present in higher dimensional fluid systems of fluid control interest. Due to the periodic boundary conditions, the KSE can be naturally expressed in terms of Fourier modes,
\begin{equation}
  u(x,t) = \sum_{k}F_k \exp\left(\dfrac{i2\pi kx}{L}\right).
 \label{eq:fouriertransform}
\end{equation}
The real-valued Fourier state space vector of the system can be described as the following, 
\begin{equation}
  F = [b_0,c_0,b_1,c_1,\dots],
 \label{eq:interleavefourier}
\end{equation}
where $F_k=b_k+ic_k$.
The dynamics of the KSE with periodic boundary conditions are equivariant under translations: i.e.~if $u(x,t)$ is a solution, then so is $u(x+\delta x,t)$ for any spatial shift $\delta x$ \citep{Budanur2017}. For an arbitrary state $F$, its translationally symmetric state differing by a phase angle of $\theta$ can be described by the following operator,
\begin{equation}
  \tau\left(\theta,F_k\right)=\exp(-ik\theta)F_k.
 \label{eq:fouriershift}
\end{equation}
Here the phase angle and spatial shift is related by $\delta x = L\theta/2\pi$. The KSE also has no preferred ``drift" direction; for each solution $u(x,t)$ is a reflection $-u(-x,t)$ that is dynamically equivalent \citep{Budanur2017}. In Fourier space, reflection symmetric states are related by a complex conjugation followed by negation, resulting in a sign change in the real component $b_k$, yielding the operator,
\begin{equation}
  \sigma(F) = [-b_0,c_0,-b_1,c_1,\dots].
 \label{eq:interleavefourierreflect}
\end{equation}

For a flow system with no preferred drift direction or spatial localization, it is natural to choose identical and uniformly spaced actuators for control. Spatially localized control is implemented in the KSE with $N=4$ equally spaced Gaussian jets located at $X\in\{0,L/4,2L/4,3L/4\}$ as done in \citep{Bucci2019}, 
\begin{equation}
  f(x,t) = \sum_{i=1}^{4}\dfrac{a(t)_i}{\sqrt{2\pi\sigma_s}}\exp\left(-\dfrac{(x-X_i)^{2}}{2\sigma_s^{2}}\right).
 \label{eq:forcingterm}
\end{equation}

To serve as an analogue to energy-saving flow control problems, we are interested in the minimization of the integral quantities of dissipation and total power input (required to power the system and jets) of the KSE system, which are described by $D = \langle u_{xx}^2\rangle$ and $P_f = \langle u_x^2\rangle+ \langle uf \rangle$, respectively. Here $\langle \cdot \rangle$ is the spatial average. 

\subsection{Deep Reinforcement Learning}

Reinforcement learning is a model-free, data-driven, method to learn the mapping function between an observed state, $s_t$, of the environment, and the action, $a_t$, that maximizes the cumulative reward, $R_t$, by experiencing the consequences of these state-action pairs. The basic RL process is cyclic: at time, $t$, the agent samples the state, $s_t$, of the environment and, in Markovian fashion, outputs an action, $a_t$, which belongs to a prescribed range of actions. This action is applied to the environment for a duration of $T=0.25$ and the environment is evolved forward in time to state, $s_{t+1}$. We note here the subscript $s_{t+1}$ is equivalent to $s_{t+T}$, but we maintain the $s_{t+1}$ nomenclature for consistency with RL literature. How desirably the environment evolved from $s_t$ to $s_{t+1}$ under the influence of action $a_t$ is then quantified by the scalar reward, $r_t$, and the process repeats. The cumulative reward, $R_t$, is the sum of discounted individual reward returns of state-action pairs,
\begin{equation}
    R_t=r_t+\gamma r_{t+1}+\gamma^2 r_{t+2}+\ldots+\gamma^{n-1} r_{n}.
   \label{eq:Rt}
\end{equation}

The discount factor, $\gamma$, is chosen to be $0<\gamma<1$, as events further into the future are more uncertain than those nearer the current instant. Here the instantaneous reward, $r_t$, computed for each observed state-action pair, was chosen to achieve our aim of minimizing the energy dissipation rate,  
\begin{equation}
  r_t = -\overline{(D + P_f)},
 \label{eq:rewardfn}
\end{equation}
where $\overline{\cdot}$ is the time average over the duration of an actuation time interval of $T$. 

\begin{figure}[t]
	\begin{center}
		\includegraphics[width=0.9\textwidth]{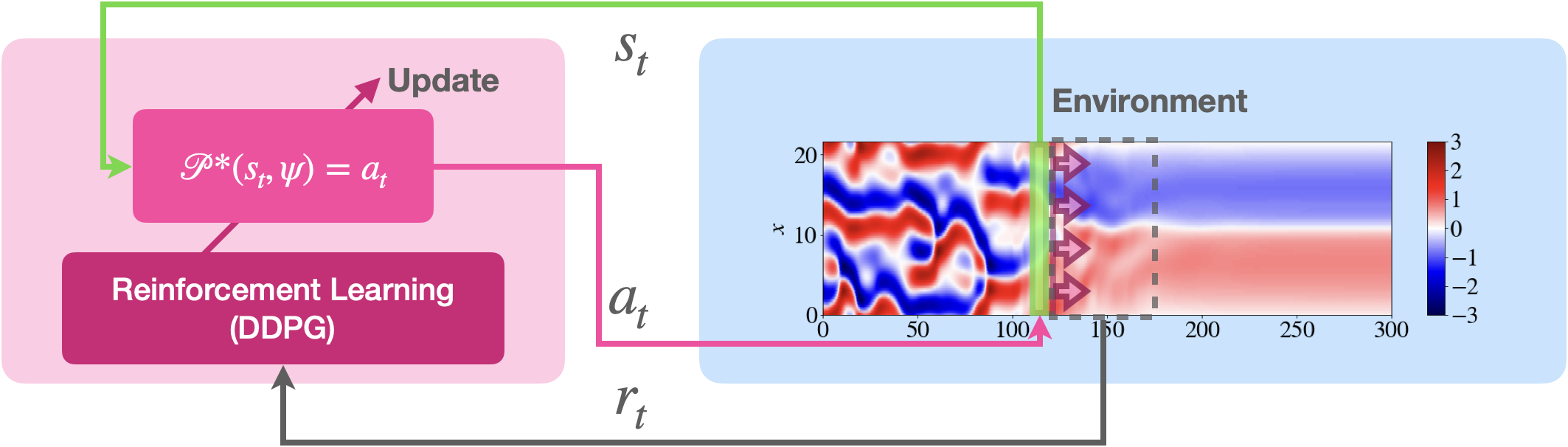}
		\caption[]{Graphical schematic of the KSE flow control problem using deep RL.}
		\label{fig:Schematic}
	\end{center}
\end{figure}

In our work the environment is the KSE, the state observation is the state of the KSE, $s_t=u(t)$, and the action output is the control signal to the Gaussian jets, $a_t=a$. An arbitrary state-action mapping function is called the policy function, $\mathscr{P}(s_t)=a_t$, whereas the optimal policy that maximizes reward is denoted as $\mathscr{P}^*$. The $\mathscr{P}^*$ learning problem is illustrated in Fig. \ref{fig:Schematic}, which shows the mapping between $s_t$ and $a_t$ to maximize $R_t$ through principles of RL which will be described later this section.

In this work we use the Deep Deterministic Policy Gradient (DDPG) algorithm \citep{Lillicrap2016}, which takes on an Actor-Critic structure and serves as the baseline deep RL algorithm from which we will introduce symmetry-reducing modifications later on. The DDPG algorithm aims to approximate two key functions with NNs: the aforementioned optimal policy, $\mathscr{P}^*$, and the optimal state-action value function, $Q^*(s,a)$. In order to understand how to learn these two functions, it is necessary to understand the $Q$ function, which quantifies the expected cumulative reward when action $a_t$ is performed on state $s_t$ given the current policy $\mathscr{P}$, 
\begin{equation}
    Q(s,a)=\mathbb{E}[R_t|s_t=s,a_t=a,\mathscr{P}].
   \label{eq:Q}
\end{equation}
We seek the policy $\mathscr{P}^*$ that yields the largest state-action value $Q^*(s,a)$. I.e.:
\begin{equation}
    Q^*(s,a)=\max_\mathscr{P} \mathbb{E}[R_t|s_t=s,a_t=a,\mathscr{P}].
   \label{eq:Qstar}
\end{equation}
Importantly, $ Q^*(s,a)$ obeys the Bellman Equation \citep{Mnih2015}:
\begin{equation}
    Q^*(s_t,a_t)=r_t+\gamma\max_{a_{t+1}}Q^*(s_{t+1},a_{t+1}).
   \label{eq:Bellman}
\end{equation}

\begin{figure}[t]
	\begin{center}
		\includegraphics[width=0.7\textwidth]{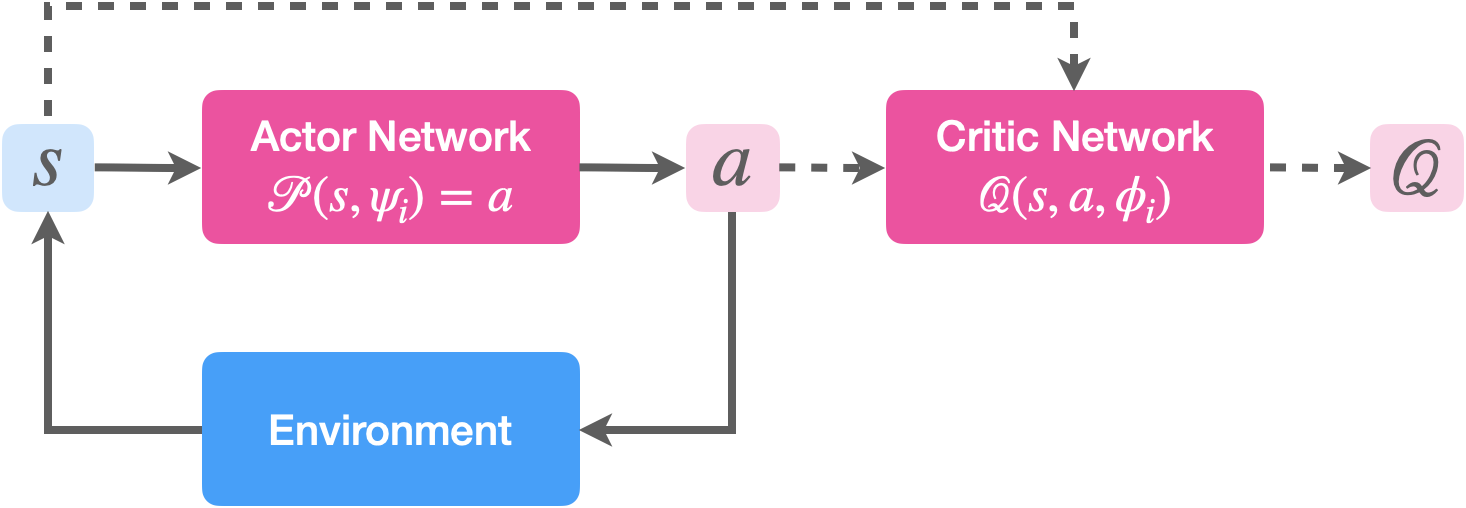}
		\caption[]{Actor-Critic learning scheme: During training the Actor and Critic Network train simultaneously (dashed, solid lines). Once training is complete, the Actor network interacts with the environment independently as the feedback controller (solid lines).}
		\label{fig:ActorCritic}
	\end{center}
\end{figure}

Obtaining $Q^*$ by explicitly evaluating all possible state-action pairs in a continuous state-action space is intractable. DDPG resolves this difficulty by utilizing NNs to approximate the $Q^*$ function and the optimal policy $\mathscr{P}^*(s)$, which are also known as the ``Critic" and ``Actor" networks, respectively \citep{Lillicrap2016}. The Critic and Actor networks are parameterized by weights $\phi$ and $\psi$, respectively,  
\begin{equation}
    Q^*(s,a)\approx Q(s,a,\phi),
   \label{eq:NN_appxQ}
\end{equation}
\begin{equation}
    \mathscr{P}^*(s)\approx \mathscr{P}(s,\psi).
   \label{eq:NN_appxP}
\end{equation}
The Actor network is generically referred to as the ``agent" in this method. Shown in Fig. \ref{fig:ActorCritic} is a schematic of the Actor-Critic learning cycle. During training, the Actor network attempts to map the state observation to the optimal action. The output action along with the state observation are then passed to the Critic network, which attempts to estimate the Q-value of the state-action pair. Note that once training is complete, the Critic network may be discarded and closed-loop control is performed between the Actor network and the environment only.

During training the weights of the Critic network are updated with the following loss function, 
\begin{equation}
    L_i(\phi_i)=[(r(s_t,a_t)+\gamma Q(s_{t+1},a_{t+1},\phi_i))-Q(s_t,a_t,\phi_i)]^2,
   \label{eq:Loss}
\end{equation}
which will be minimized when the Bellman Equation, Eq. \ref{eq:Bellman}, is satisfied, signaling that the optimal policy has been approximated. This loss is used for back-propagation through the Critic network and the resulting gradient is utilized in updating the Actor network \citep{Lillicrap2016}. The optimization algorithms implemented in training assume that samples are distributed independently and identically, which is generally untrue for data generated from our exploratory trajectories. To mitigate this, the algorithm is trained on minibatches of experience tuples, $e_t=(s_t,a_t,r_t,s_{t+1})$, selected randomly from a memory cache of past experience tuples. This memory cache technique is called \textit{experience replay} and is implemented to combat the instabilities in $Q$-learning caused by highly correlated training sets \citep{Mnih2015}. The DDPG algorithm used is shown in Algorithm \ref{ALG:DDPG}.

\begin{algorithm}[t]
    \SetAlgoLined
     Initialize Critic network $Q(s,a,\phi)$, Actor network $\mathscr{P}(s,\psi)$ with random weights $\phi$, $\psi$;
     Initialize target networks $Q',\mathscr{P}'$ with weights $\phi'\leftarrow \phi$, $\psi'\leftarrow\psi$\;
     Initialize $\beta$ = 1.0 \;
     \For{episode = 1, M}{
      Initialize $s_1$ \;
      Initialize OU process, $\mathscr{N}$, for action exploration\;
      \For{t = 1, T}{
       Select action $a_t = \mathscr{P}(s_t,\psi) + \beta\mathscr{N}_t$\;
       apply action $a_t$ to environment, observe $r_t$, $s_{t+1}$ \;
       store memory tuple ($s_t,a_t,r_t,s_{t+1}$) in replay memory cache
       update $s_t=s_{t+1}$\;
       sample random minibatch of $e_j$=($s_j,a_j,r_j,s_{j+1}$) from mem. cache\;
       set $y_j=r_j + \gamma Q'(s_{j+1},a_{j+1},\phi')$ \;
       Update critic network: $L =\frac{1}{N}\sum_j (y_j-Q(s_j,a_j,\phi))^2$, Eq. \ref{eq:Loss} \;
       Update actor network with learning rate $\alpha_\psi$: $\psi\leftarrow\psi+\alpha_\psi\frac{1}{N}\sum_j \nabla_a Q(s_i,a_i,\phi) \nabla_\psi \mathscr{P}(s_i,\psi)$ \citep{Lillicrap2016} \;
       update target networks with learning rate $\alpha_T$: $\phi'\leftarrow \alpha_T\phi + (1-\alpha_T)\phi'$, $\psi' \leftarrow \alpha_T\psi+ (1-\alpha_T)\psi'$\;
       $\beta$ $\leftarrow$ $D_{\beta}$ $\beta$\;
       }{
      }
     }
    \caption{Deep Deterministic Policy Gradient with Experience Replay}
    \label{ALG:DDPG}
\end{algorithm}

We utilize Actor-Critic networks each with two hidden layers of size 256 and 128 with ReLU activation functions. The output layer of the Actor and Critic networks are composed of tanh and linear activation functions, respectively. Increasing the hidden layer size did not appear to strongly influence overall performance. We employ a rolling experience replay buffer of generated training data, $(s_t,a_t,r_t,s_{t+1})$, of size 500,000 experiences and update with batch sizes of 128. We trained each agent for 4000 episodes, with each episode initialized randomly on the unforced KSE attractor and lasting 100 time units (400 actions). For exploration of state-action space during training we employ Ornstein-Uhlenbeck noise, $\mathscr{N}$, \citep{Lillicrap2016} to encourage action exploration.


\subsection{State-Action Space Symmetries and the Deep RL Problem}

It is important to now consider the symmetry of the overall controlled system. As mentioned earlier, the KSE possesses a continuous translational symmetry and a discrete reflection symmetry. The $N$ equidistant and spatially-fixed actuator jets also possess their own symmetries of a $N$ discrete translational-shift and reflection symmetry. The symmetry of the overall controlled system is the intersection of the symmetry group operations, which in this problem is simply a $N$ discrete translational-shift, $\tau_N$, and reflection symmetry. What this indicates is that a dynamically equivalent state-action pair, which can be thought of as a state relative to the locations of the jets, can appear in $2N$ different orientations ($N$ discrete translations, and their $N$ respective reflections), dividing state-action space into $2N$ dynamically equivalent sectors. The impact this degeneracy has on learning the optimal policy is twofold. First, the $\mathscr{P}$ function should map dynamically equivalent states to dynamically equivalent actions, which requires the Actor network to learn to be an equivariant function with respect to the total controlled system's symmetries, e.g. $\tau_N(\mathscr{P}(s,\psi))=(\mathscr{P}(\tau_N(s),\psi))$. Second, the $Q$ function should be invariant to discrete translations and reflections of the state-action input, which requires the Critic network to learn to be an invariant function of dynamically equivalent state-action pairs, e.g. $Q(s,a,\phi)=Q(\sigma(s),\sigma(a),\phi)$.

\begin{figure}[t]
 	\centering
	\begin{subfigure}[t]{0.95\textwidth}
		\includegraphics[width=\textwidth]{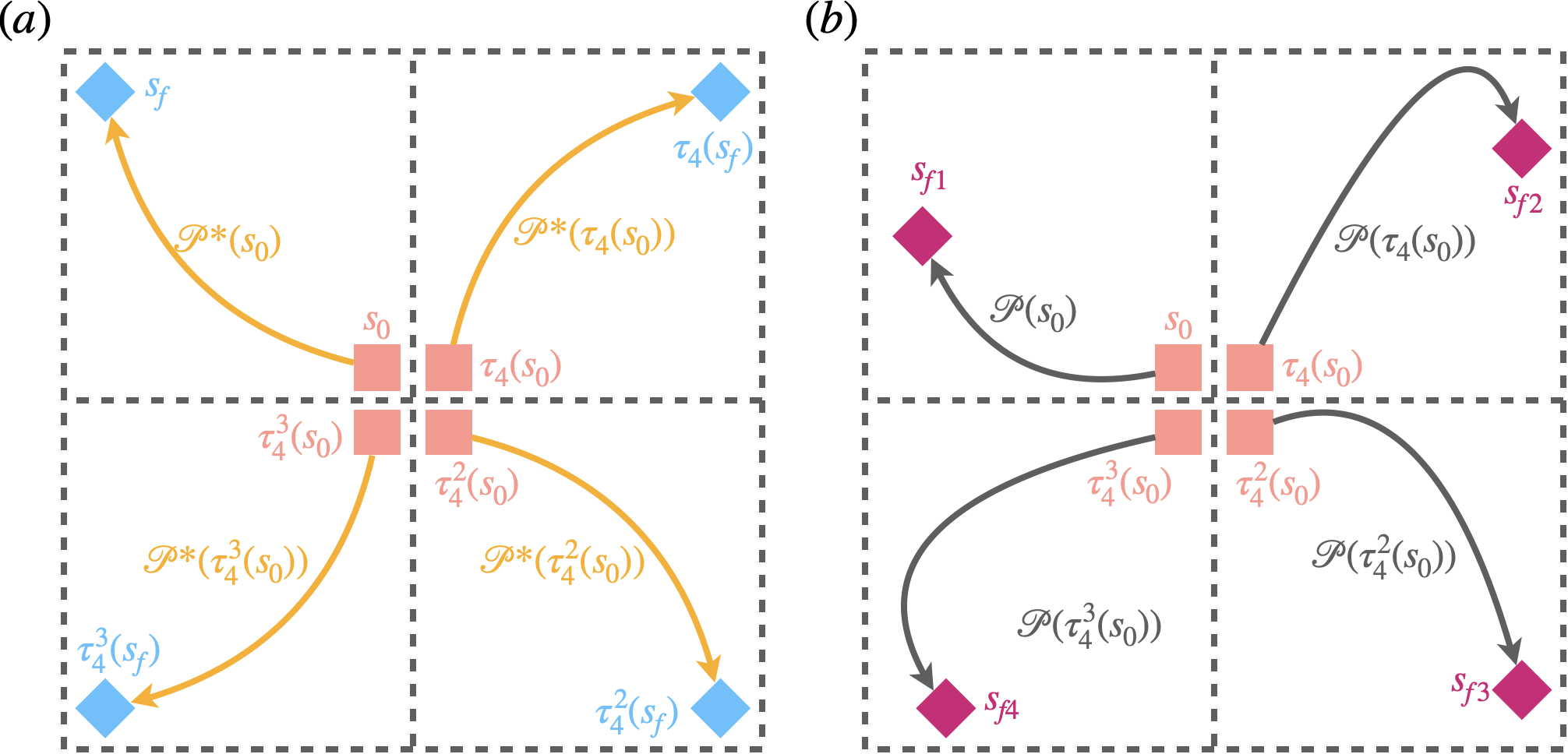}
		\phantomcaption
		\label{fig:SymmetryGraphic-a}
	\end{subfigure}
	\begin{subfigure}[t]{0.0\textwidth}
		\includegraphics[width=\textwidth]{figures/SymmetryGraphic}
		\phantomcaption
		\label{fig:SymmetryGraphic-b}
	\end{subfigure}
	\caption[]{(a) For the initial condition $s_0$, the optimal controlled trajectory following the optimal policy $\mathscr{P}^*$ leads to state $s_f$. For simplicity, $\tau_4(s)=\tau_4(\pi/2,s)$. Given a translation of the $s_0$ by $\tau_4$, following the optimal policy $\mathscr{P}^*(s)$ should yield a dynamically equivalent trajectory translated by $\pi/2$. (b) A policy that does not have symmetry enforced, $\mathscr{P}(s)$, yielding dynamically nonequivalent trajectories.}
	\label{fig:SymmetryGraphic-full}
\end{figure}

Because NNs are not intrinsically equivariant nor invariant, naive NNs must learn and internally approximate the optimal policy $2N$ times, one for each of the $2N$ dynamically equivalent sectors of state-action space. The implications of this are twofold. First, this requirement to learn redundant policies within the same network can be viewed as an implicit weight constraint that not only exhausts network capacity, but also requires an ergodic exploration of all of state-action space to generate sufficient training data. Inaccuracies in properly estimating the $Q$ function lead to poor approximations in the optimal policy, and ultimately poor training and control performance. Second, in tow with control performance, if the policy approximation is unable to map dynamically equivalent states to dynamically equivalent actions, the policy cannot be the optimal policy. An illustration of this in a cartoon-world example is shown in Fig. \ref{fig:SymmetryGraphic-full}, which possess a $\tau_4$ (discrete translational shift) symmetry. Note that given dynamically equivalent states, the optimal policy, shown in Fig. \ref{fig:SymmetryGraphic-a}, should produce dynamically equivalent trajectories as a consequence of producing dynamically equivalent actions. A sub-optimal approximation of the optimal policy that does not respect symmetry, shown in Fig. \ref{fig:SymmetryGraphic-b}, yields dynamically nonequivalent actions, and thus dynamically nonequivalent trajectories and final states despite being given dynamically equivalent initial conditions.





Our method to circumvent these limitations is to move the deep learning problem to a symmetry-reduced subspace. We accomplish this by reducing the translational symmetry of the state observation via a modification of the method of slices \citep{Budanur2017}, followed by a reflection symmetry reduction operation in order to obtain a discrete translational and reflection-reduced state. This symmetry-reduced state is then passed to the agent, which then outputs the corresponding optimal symmetry-reduced action. The previously removed symmetries are then reintroduced to the output actions prior to implementation to ensure that they respect the true orientation of the system.

We first reduce discrete translational symmetries with a modified method of slices. This operation moves all state observations to the same discrete reference phase while preserving the relative location of the $N$ actuators to the original and discrete translation-reduced state. The phase angle of the state can be calculated via Eq.(\ref{eq:phaseangle}), 
\begin{equation}
  \theta_1 = \text{arctan2}\left(b_1,c_1\right),
 \label{eq:phaseangle}
\end{equation}
where $\text{arctan2}(b,c)$, not to be confused with $\text{arctan}^2(b)$, is the 2-argument arctangent function that returns the phase of a complex number and is bounded by $-\pi$ and $\pi$. To preserve the uniqueness of a state-action pair, i.e. the relative location of the state to the $N$ spatially fixed actuators, the state phase angle, $\theta_1$, is rounded up to the nearest discrete phase, $\theta_N$,
\begin{equation}
  \theta_N=\dfrac{2\pi}{N}\text{ceil}\left(\dfrac{\theta_1}{2\pi/N}\right).
 \label{eq:discreteangle}
\end{equation}
The Fourier state, $F$, is moved into the discrete translation-reduced subspace via the discrete translational reduction operator, $\hat{F}_k=\tau_N\left(\theta_N,F_k\right)$, where
\begin{equation}
  \tau_N\left(\theta_N,F_k\right)=\exp(ik\theta_N)F_k.
 \label{eq:disNfouriershift}
\end{equation}
In the discrete translation-reduced subspace, $\hat{F}$ preserves the relative location with respect to the actuators. The resulting real-valued Fourier state space vector in the discrete translation-reduced subspace is then,
\begin{equation}
  \hat{F} = \left[\hat{b}_0,\hat{c}_0,\hat{b}_1,\hat{c}_1,\hat{b}_2,\hat{c}_2,\hat{b}_3,\hat{c}_3,\dots \right].
 \label{eq:TransRealVector}
\end{equation}

Within the discrete translation-reduced subspace, reflection symmetric states are related by the following reflection operator with respect to $N$,
\begin{equation}
  \sigma_N(\hat{F}) = \exp\left(\dfrac{2\pi}{N}ik\right)\sigma(\hat{F}).
 \label{eq:GeneralSigmaN}
\end{equation}

For $N=4$, the discrete translation-reduced reflection operator is defined as the following repeating sequence,
\begin{equation}
  \sigma_4(\hat{F}) = \left[-\hat{b}_0,\hat{c}_0,-\hat{c}_1,-\hat{b}_1,\hat{b}_2,-\hat{c}_2,\hat{c}_3,\hat{b}_3,\dots \right].
 \label{eq:FlipRealVector}
\end{equation}

We note that the sign of $\hat{c}_2$ is the first unique value that can distinguish between two reflection symmetric states within the discrete translation-reduced subspace and thus construct a reflection indicator function $\rho=\text{sign}(\hat{c}_2)$. The reflection operator, $\sigma_4$, is then applied if the indicator value $\rho<0$ to collapse reflection symmetric states into a common half of the discrete translation-reduced subspace:
\begin{equation}
  \breve{\hat{F}}=\begin{cases}
    \sigma_4(\hat{F}), & \text{if $\rho<0$}.\\
    \hat{F}, & \text{otherwise}.
  \end{cases}
  \label{eq:flipcondition}
\end{equation}

\begin{figure}[t]
	\begin{center}
		\includegraphics[width=0.9\textwidth]{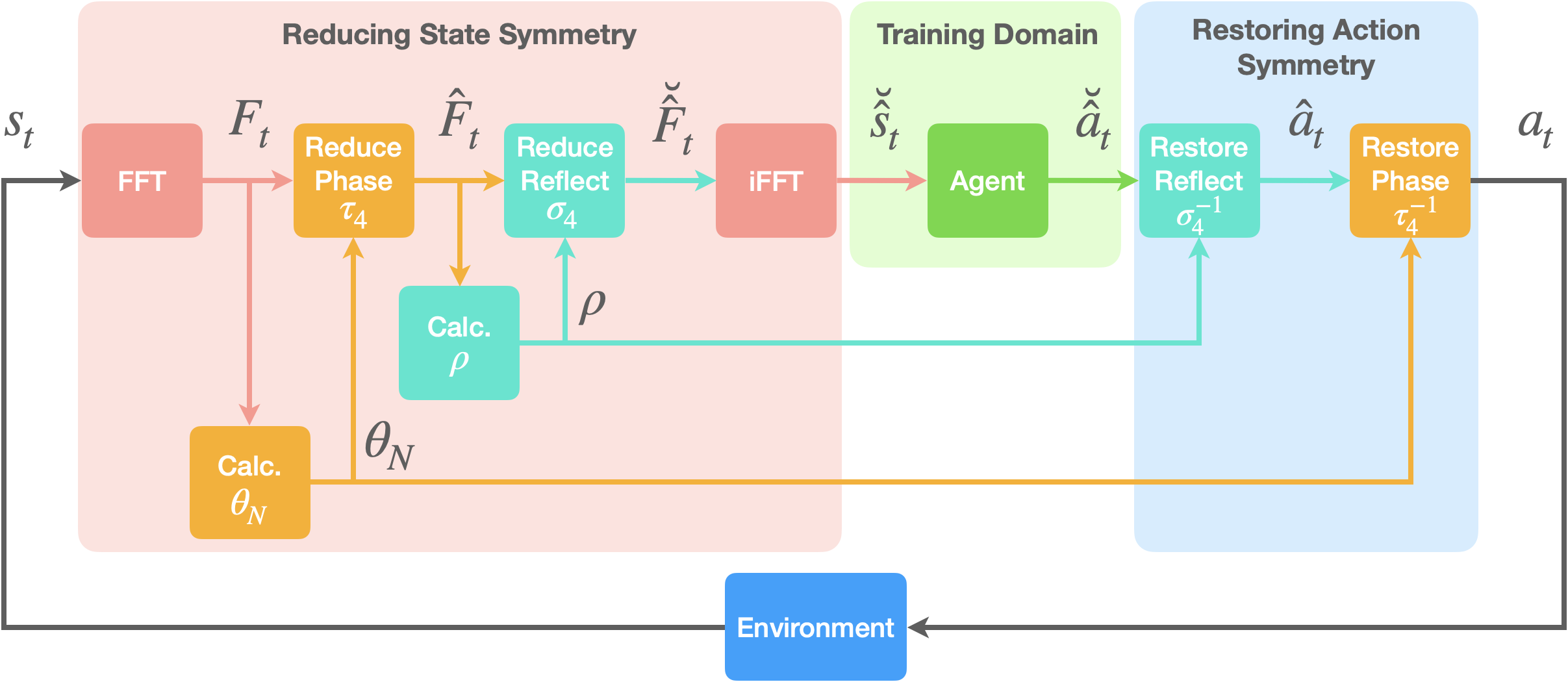}
		\caption[]{Flow diagram of discrete-symmetry reduced deep reinforcement learning. State observations have discrete translational symmetries reduced by $\tau_4$, followed by a reflection symmetry reduction by $\sigma_4$. The symmetry reduced state is then passed to the agent, which outputs a symmetry reduced action. The previously removed reflection and translational symmetries are reintroduced to the output action before being implemented in the environment.}
		\label{fig:DiscreteSymmetryReductionDRL2}
	\end{center}
\end{figure}

The resulting discrete translational and reflection-reduced Fourier state, $\breve{\hat{F}}$, is then Fourier transformed back to the real domain and passed to the deep RL agent as the state observation. A schematic of this symmetry reduction process is shown in Fig. \ref{fig:DiscreteSymmetryReductionDRL2}. By performing the appropriate transformationss to the state, the deep RL agent learns purely within a discrete symmetry-reduced subspace, and thus outputs discrete symmetry-reduced actions, $\breve{\hat{a}}$. As a result, the agent trains with symmetry-reduced experiences, $\breve{\hat{e}}_t=(\breve{\hat{s}}_t,\breve{\hat{a}}_t,r_t,\breve{\hat{s}}_{t+1})$. Importantly, because the output actions are symmetry-reduced actions, we must ensure they respect the orientation of the true state of the system before we apply them to the environment. This requires that the reverse symmetry operations that were applied to the state be applied to the actions prior to actuation. The control signal generated by the agent is therefore reflected if $\rho<0$, then rotated by $N\theta_N/2\pi$, before being applied to the system. This ensures that dynamically equivalent states receive dynamically equivalent actions. Note that in this work the Actor-Critic deep RL model is inserted in ``training domain", but generally any deep RL model may be chosen. The symmetry-reduced DDPG algorithm is shown in Algorithm \ref{ALG:SRDDPG}.

\begin{algorithm}[t]
    \SetAlgoLined
     Initialize Critic network $Q(s,a,\phi)$, Actor network $\mathscr{P}(s,\psi)$ with random weights $\phi$, $\psi$;
     Initialize target networks $Q',\mathscr{P}'$ with weights $\phi'\leftarrow \phi$, $\psi'\leftarrow\psi$\;
     Initialize $\beta$ = 1.0 \;
     \For{episode = 1, M}{
      Initialize $s_1$ and compute $\breve{\hat{s}}_{1}$\;
      Initialize OU process, $\mathscr{N}$, for action exploration\;
      \For{t = 1, T}{
       Select action $\breve{\hat{a}}_{t} = \mathscr{P}(\breve{\hat{s}}_{t},\psi) + \beta\mathscr{N}_t$, compute $a_t$\;
       Apply action $a_t$ to environment, observe $r_t$, $s_{t+1}$\;
       Compute $\breve{\hat{s}}_{t+1}$\;
       Store symmetry reduced memory tuple ($\breve{\hat{s}}_{t},\breve{\hat{a}}_{t},r_t,\breve{\hat{s}}_{t+1}$) in mem. cache\;
       update $s_t=s_{t+1}$, $\breve{\hat{s}}_{t}=\breve{\hat{s}}_{t+1}$\;
       sample random minibatch of $e_j$=($\breve{\hat{s}}_{j},\breve{\hat{a}}_{j},r_j,\breve{\hat{s}}_{j+1}$) from mem. cache\;
       set $y_j=r_j + \gamma Q'(\breve{\hat{s}}_{j+1},\breve{\hat{a}}_{j+1},\phi')$ \;
       Update critic network: $L =\frac{1}{N}\sum_j (y_j-Q(\breve{\hat{s}}_{j},\breve{\hat{a}}_{j},\phi))^2$, Eq. \ref{eq:Loss} \;
       Update actor network with learning rate $\alpha_\psi$: $\psi\leftarrow\psi+\alpha_\psi\frac{1}{N}\sum_j \nabla_a Q(\breve{\hat{s}}_{j},\breve{\hat{a}}_{j},\phi) \nabla_\psi \mathscr{P}(\breve{\hat{s}}_{j},\psi)$ \citep{Lillicrap2016} \;
       update target networks with learning rate $\alpha_T$: $\phi'\leftarrow \alpha_T\phi + (1-\alpha_T)\phi'$, $\psi' \leftarrow \alpha_T\psi+ (1-\alpha_T)\psi'$\;
       $\beta$ $\leftarrow$ $D_{\beta}$ $\beta$\;
       }{
      }
     }
    \caption{Deep Deterministic Policy Gradient with Symmetry Reduction}
    \label{ALG:SRDDPG}
\end{algorithm}

\section{Results} \label{Results}

Section \ref{sec:Performance} presents a quantitative comparison between a naive (i.e.~``symmetry-unaware") agent, a naive agent trained with augmented data, and a symmetry-reduced agent. The learned control strategy and its dynamical significance are characterized in Section \ref{sec:Characterization} while Section \ref{sec:LQR}  examines a classical LQR approach to control for this problem. Comparison with the RL results provides some insight into why the RL algorithm learns the policy that it did.  Finally, in Section \ref{sec:Robustness}, the robustness of the agent to input-output noise and perturbations to system parameters is evaluated.


\subsection{Performance Comparison} \label{sec:Performance}
\begin{figure}[t]
 	\centering
	\begin{subfigure}[t]{0.95\textwidth}
		\includegraphics[width=\textwidth]{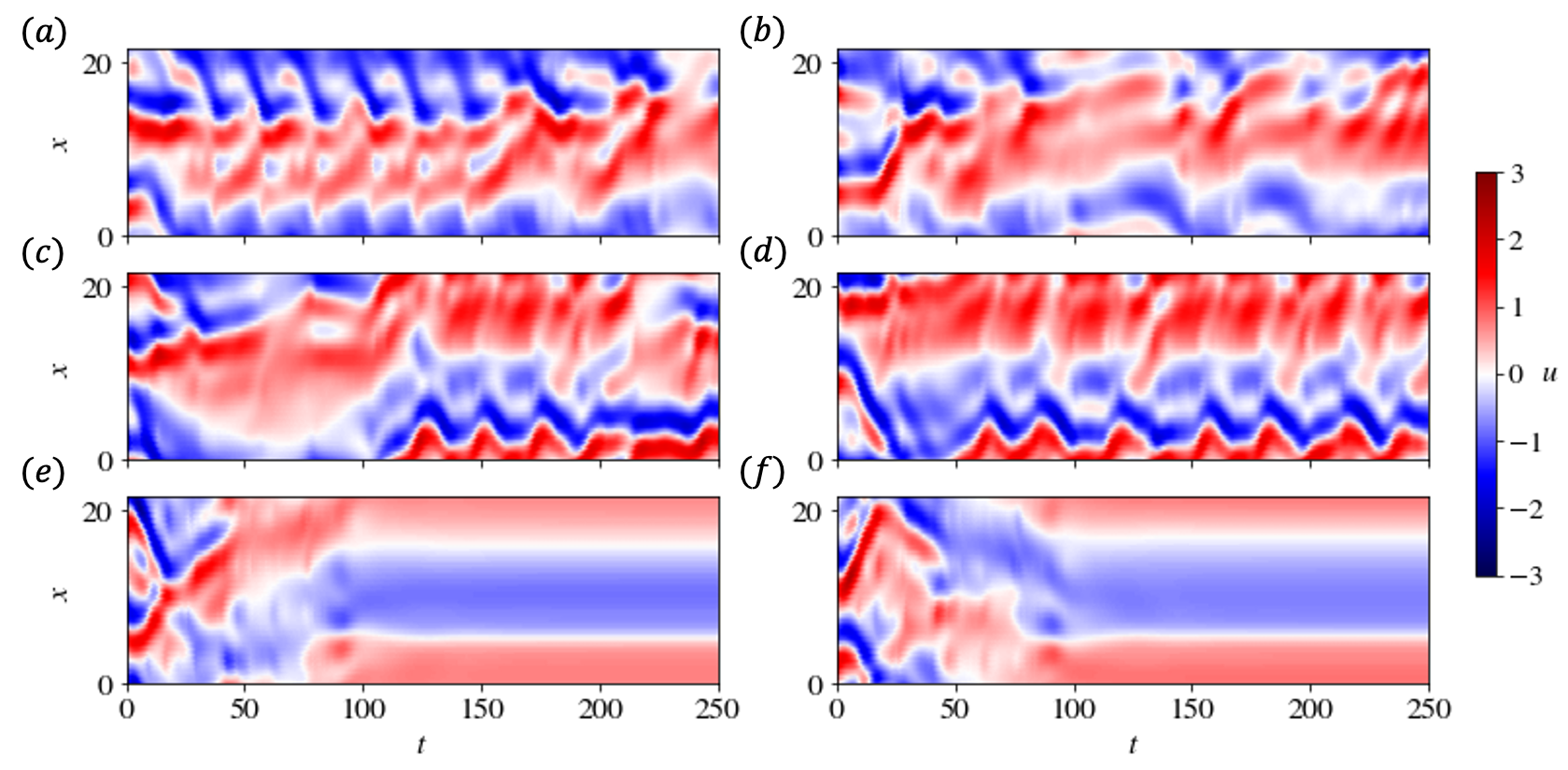}
		\phantomcaption
		\label{fig:ExampleSymmetricTrajectory-a}
	\end{subfigure}
	\begin{subfigure}[t]{0.0\textwidth}
		\includegraphics[width=\textwidth]{figures/SymmetricTraj}
		\phantomcaption
		\label{fig:ExampleSymmetricTrajectory-b}
	\end{subfigure}
	\begin{subfigure}[t]{0.0\textwidth}
		\includegraphics[width=\textwidth]{figures/SymmetricTraj}
		\phantomcaption
		\label{fig:ExampleSymmetricTrajectory-c}
	\end{subfigure}
	\begin{subfigure}[t]{0.0\textwidth}
		\includegraphics[width=\textwidth]{figures/SymmetricTraj}
		\phantomcaption
		\label{fig:ExampleSymmetricTrajectory-d}
	\end{subfigure}
	\begin{subfigure}[t]{0.0\textwidth}
		\includegraphics[width=\textwidth]{figures/SymmetricTraj}
		\phantomcaption
		\label{fig:ExampleSymmetricTrajectory-e}
	\end{subfigure}
	\begin{subfigure}[t]{0.0\textwidth}
		\includegraphics[width=\textwidth]{figures/SymmetricTraj}
		\phantomcaption
		\label{fig:ExampleSymmetricTrajectory-f}
	\end{subfigure}
	\caption[]{Pairs of controlled trajectories with dynamically equivalent initial conditions (translated by $L/2$ and reflected) controlled by the (a),(b) Naive agent, (c),(d) augmented naive agent, and (e),(f) symmetry-reduced agent.}
	\label{fig:ExampleSymmetricTrajectory-full}
\end{figure}

To illustrate the importance of discrete symmetry reduction, the naive and symmetry-reduced agents are tested with dynamically equivalent initial conditions, which are related by a translation of half the domain and a reflection operation. 
The resulting trajectories controlled by the naive agent are shown in Fig. \ref{fig:ExampleSymmetricTrajectory-a}, which demonstrate the naive agent's inability to map dynamically equivalent states to dynamically equivalent actions. The dynamically inequivalent trajectories are a product of the naive agent being unable to map dynamically equivalent states to dynamically equivalent actions. This is due to the inherent difficulty for the NNs to consolidate and learn identical optimal sub-policies for each symmetric sector of state-action space (i.e. it is unable to become equivariant to symmetry-related state-action pairs). 
 
  To aid the implicit learning of dynamically equivalent state-action mappings, we also trained naive agents with additional synthetic training data, produced by applying symmetry operations to the originally generated data. Agents trained with this augmented data set will be called augmented naive agents. Shown in Fig. \ref{fig:ExampleSymmetricTrajectory-c} and Fig. \ref{fig:ExampleSymmetricTrajectory-d} are trajectories controlled by an augmented naive agent beginning with dynamically equivalent initial conditions. This augmented naive agent also produces dynamically nonequivalent trajectories, which indicate that it fails to learn dynamically equivalent state-action mappings.
  
  In contrast, the symmetry-reduced agent, given dynamically equivalent initial conditions, will produce dynamically equivalent actions and therefore dynamically equivalent controlled trajectories, which are shown in Fig. \ref{fig:ExampleSymmetricTrajectory-e} and Fig. \ref{fig:ExampleSymmetricTrajectory-f}. As a result, the symmetry-reduced method inherently yields improved performance variance and robustness to initial conditions over the naive method, as the controlled trajectories of the naive agent depends on the orientation of the state while the symmetry reduced method does not. Furthermore, while the dynamics of the naively controlled system remain chaotic, those with the symmetry-aware controller evolve to a low-dissipation steady state, a phenomenon that we analyze further in Section \ref{sec:Characterization}.

\begin{figure}[t]
	\begin{center}
		\includegraphics[width=0.9\textwidth]{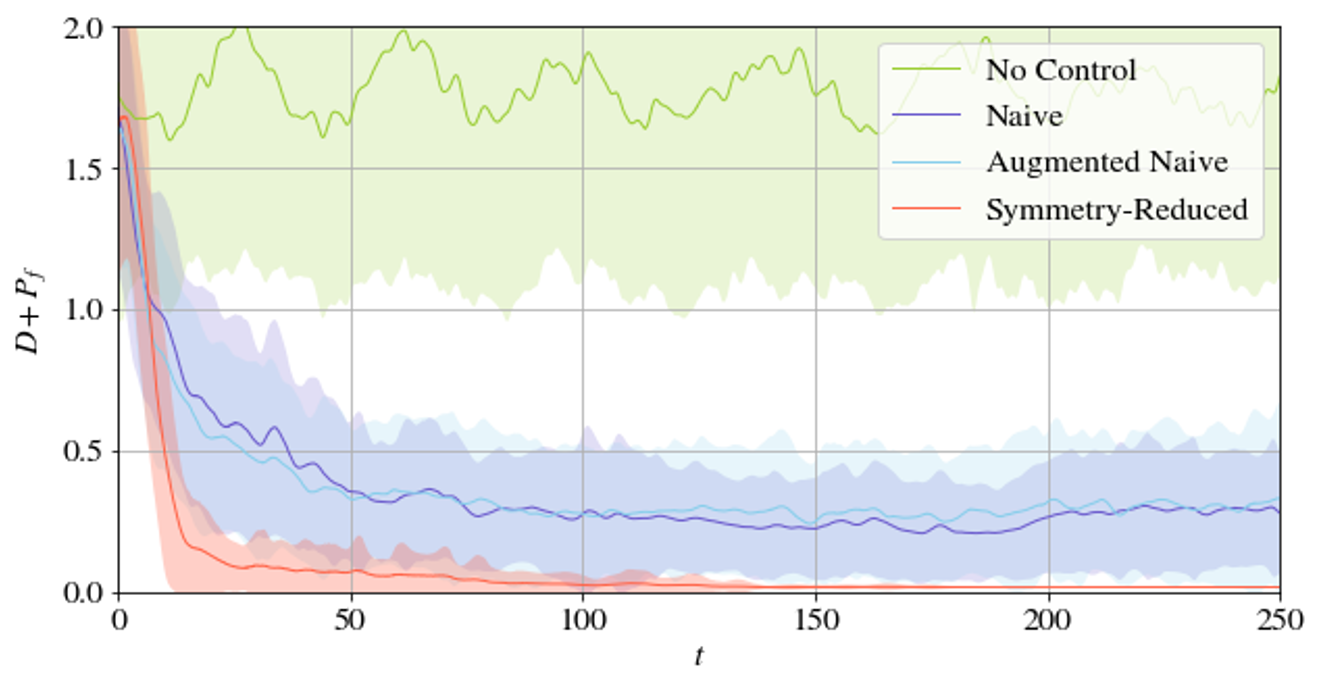}
		\caption[]{Ensemble mean $D+P_f$ of 100 trajectories controlled by: No control (green), Naive (purple), Augmented naive (blue), and translation+reflection-reduced (red). Each initial condition is randomly initialized on the KSE attractor. Standard deviation of each ensemble is shaded in its respective color.}
		\label{fig:EnsembleDissipation}
	\end{center}
\end{figure}

To demonstrate this improved performance, the naive and symmetry-reduced agent are tested to control 100 random initial conditions sampled from the attractor of the unforced KSE. The controlled trajectory duration is extended to 250 time units, 2.5 times the 100 time unit duration experienced during training, to examine the robustness of the policies beyond the training time horizon. The mean and standard deviation of $D+P_f$ of the 100 controlled trajectories are shown in Fig. \ref{fig:EnsembleDissipation} with respect to time. Notably, the addition of symmetry-reduction yields controlled trajectories with significantly lower mean $D+P_f$ and tighter variance than compared to the naive and augmented agents. Furthermore, the symmetry-reduced agent reaches its low $D+P_f$ target state in a much shorter time than the naive agents.

\begin{figure}[t]
	\begin{center}
		\includegraphics[width=0.9\textwidth]{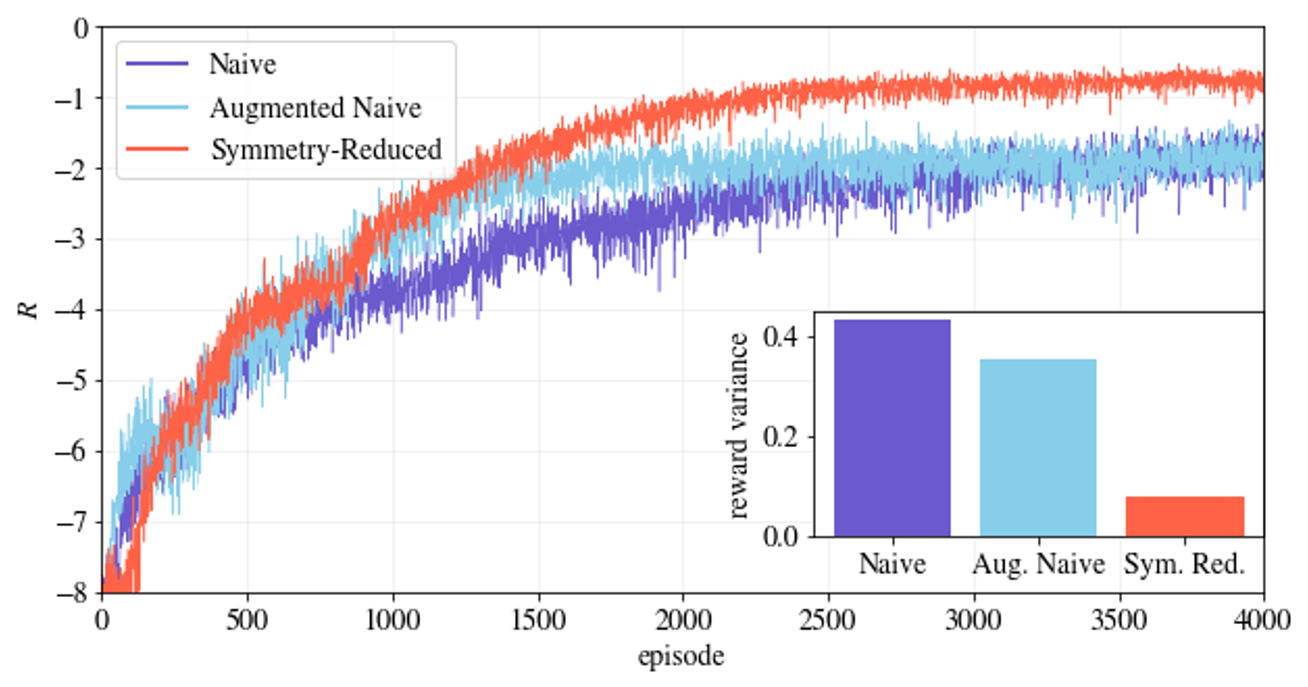}
		\caption[]{Ensemble training reward (Will change to be consistent with introduction nomenclature) (10 models each) vs. episode for: Naive (purple), Augmented Naive with synthetic symmetric data (blue), and translation+reflection-reduced (red). The reward variance of the last 1000 episodes is shown in the inset.}
		\label{fig:EnsembleTraining}
	\end{center}
\end{figure}

The advantage of symmetry-reducing the RL problem also appears in training. In Fig. \ref{fig:EnsembleTraining}, the mean reward return of 10 models is shown for each RL method with respect to training episode. The symmetry-reduced agents not only reaches greater reward returns than compared to the naive agents given the same amount of training, but they do so in significantly fewer training episodes, demonstrating the enhanced efficiency in training data usage. This improved training efficiency is a result of the symmetry-reduced agents only needing to learn one symmetric sector of state-action space, as opposed to the naive agents, which must rely completely on ergodicity to explore and learn all of state-action space. Furthermore, the variance in the training reward-return of the symmetry-reduced agents are much lower than that of the naive agents. This also highlights the improved policy robustness, as each episode differs in initial conditions and noisy actuation perturbations (exploration noise).

 We also comment that the augmented naive agents with the additional synthetic symmetric training data exhibited improved early training performance compared to the naive agents. However, at the end of training, the augmented naive agents ultimately achieve similar reward returns as the naive agents without synthetic training data. 

\subsection{Characterizing the Learned Control Solution} \label{sec:Characterization}
\begin{figure}[t]
 	\centering
	\begin{subfigure}[t]{0.95\textwidth}
		\includegraphics[width=\textwidth]{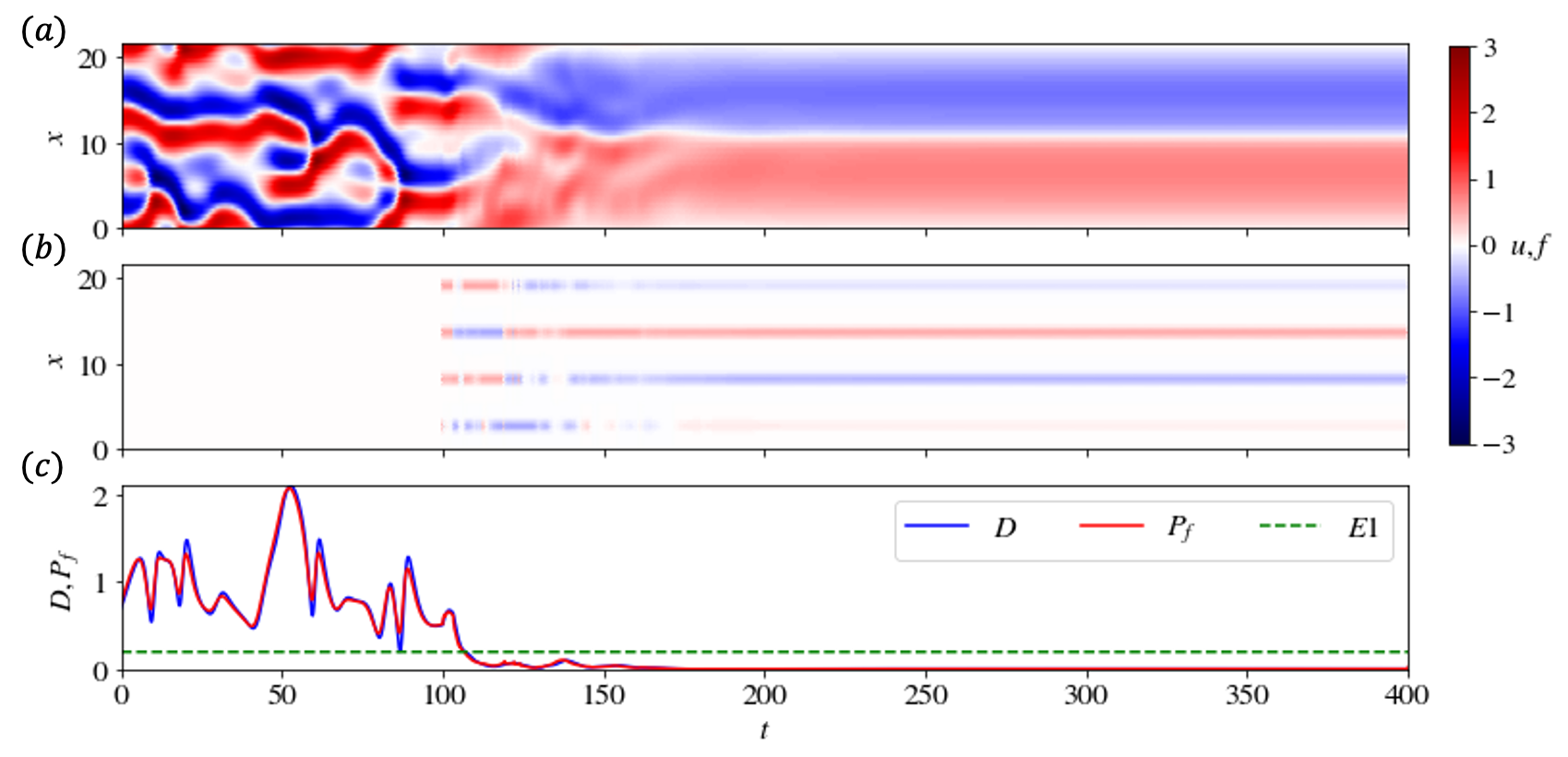}
		\phantomcaption
		\label{fig:ControlledTrajectory-a}
	\end{subfigure}
	\begin{subfigure}[t]{0.0\textwidth}
		\includegraphics[width=\textwidth]{figures/ControlledTraj}
		\phantomcaption
		\label{fig:ControlledTrajectory-b}
	\end{subfigure}
	\begin{subfigure}[t]{0.0\textwidth}
		\includegraphics[width=\textwidth]{figures/ControlledTraj}
		\phantomcaption
		\label{fig:ControlledTrajectory-c}
	\end{subfigure}
	\caption[]{(a) Symmetry-reduced agent controlled trajectory vs. time. The controller is turned on at $t=100$. (b) Forcing profile vs. time. (c) The dissipation and total power cost as a function of time. For reference the dissipation of $E1$ is included.}
	\label{fig:ControlledTrajectory-full}
\end{figure}

We noted above that the symmetry-aware agent drives the system from chaotic dynamics to a steady state. Fig.~\ref{fig:ControlledTrajectory-a} illustrates such a controlled trajectory, along with the control action $f(t)$ in Fig.~\ref{fig:ControlledTrajectory-b}, and the $D(t)$, $P_f(t)$ in Fig.~\ref{fig:ControlledTrajectory-c}. In this figure, control action begins at $t=100$.  Qualitatively, the control action occurs in two phases. The first phase, approximately $t=100$ to $t=180$, is characterized by complex transient actuations that navigate the system to the neighborhood of a steady state. The second phase, approximately $t=180$ and onward, is characterized by an essentially constant forcing profile with an extremely small time dependent residual corresponding with stabilizing the equilibrium state. We denote this ``constant" forcing as $f=\alpha_{22}$, where $\alpha$ denotes the forcing profile and the subscript corresponds to the respective domain length $L$, and the steady state as $u_{\alpha_{22}}$. When control is removed the system returns to its original chaotic dynamics. 

\begin{figure}[t]
 	\centering
	\begin{subfigure}[t]{0.95\textwidth}
		\includegraphics[width=\textwidth]{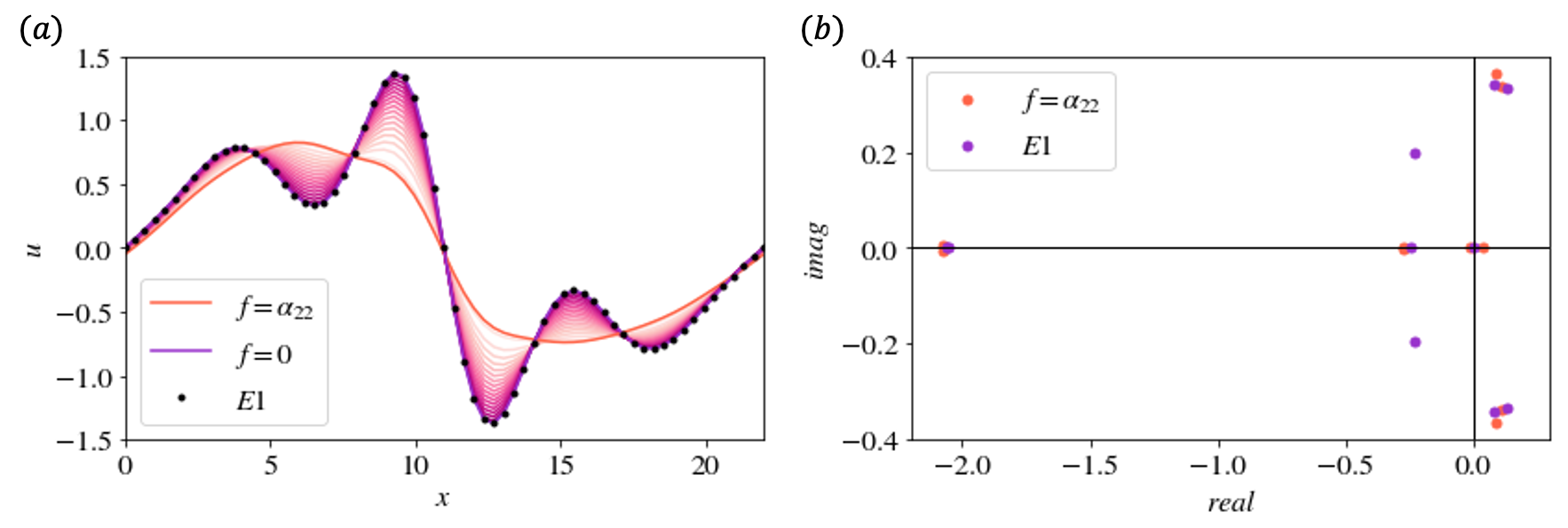}
		\phantomcaption
		\label{fig:FContinuationL22-a}
	\end{subfigure}
	\begin{subfigure}[t]{0.0\textwidth}
		\includegraphics[width=\textwidth]{figures/L22ContandEigen}
		\phantomcaption
		\label{fig:FContinuationL22-b}
	\end{subfigure}

	\caption[]{(a) Force-continuation solutions from $f=\alpha_{22}$ (red) to $f=0$ (purple), yielding equilibrium solutions from $E_{\alpha_{22}}$ to $E1$ (dots). (b) The leading eigenvalues of the KSE linearized around $E_{\alpha_{22}}$ and $E1$.}
	\label{fig:FContinuationL22-full}
\end{figure}

To understand the appearance of this steady state  in the controlled system, we first note that at $L=22$ in the absence of control, the KSE has a number of unstable steady states, some of which lie in the vicinity of the chaotic attractor \citep{Cvitanovic2010}.
To investigate the dynamical connection between the state $u_{\alpha 22}$ found with the RL agent with forcing profile $f=\alpha_{22}$ and the uncontrolled KSE, continuation was performed between the final forced system to the unforced system. We accomplished our forcing continuation by iteratively Newton-solving for an equilibrium solution that satisfies Eq. (\ref{eq:KSE}) with initial solution guess of $u_{\alpha 22}$ and forcing profile $f_0=\alpha_{22}$. The resulting solution, $E_{\alpha_{22}}$, was then used as an initial guess for a system with slightly reduced forcing amplitude. The process was repeated until $f=0$, where the solution converged to an equilibrium solution of the unforced KSE. The solutions found for incremental scalings of $f=\alpha_{22}$ are shown in Fig. \ref{fig:FContinuationL22-a}, which converge to a known solution of the KSE denoted by \citep{Cvitanovic2010} as $E1$. Interestingly, $E1$ is the lowest-dissipation solution known aside from the trivial zero solution, and  $E_{\alpha_{22}}$ exhibits even lower dissipation and power input than  $E1$. The discovered $E_{\alpha_{22}}$ solution corresponds to the $E1$ solution modified by the ``jets" in a manner that smooths its peaks, leading to weaker gradients and thus lower dissipation.

Shown in Fig. \ref{fig:FContinuationL22-b} are the leading eigenvalues of the $E1$ and $E_{\alpha_{22}}$ solutions, demonstrating that they are both linearly unstable. As the $E_{\alpha_{22}}$ solution is linearly unstable, we note that the agent maintains this state with oscillatory-like adjustments that are several orders of magnitude smaller than the mean actuation about $f=\alpha_{22}$. 


Importantly, nowhere in the algorithm was $E_{\alpha_{22}}$ explicitly targeted -- the algorithm discovers and stabilizes an underlying unstable steady state of the dynamical system, despite having been given no information about such solutions. Stabilizing an underlying steady state of the dynamical system is an efficient strategy as it requires less control effort than brute-forcing the system to a region of state space where it would not naturally reside. We speculate that this ``strategy" of finding and stabilizing an unstable recurrent solution (steady state, periodic orbit) might arise in RL control of a wide variety of systems displaying complex dynamics. We contrast this with other recent data-driven control-target identifying methods, such as \citep{Bramburger2020} which identifies and stabilizes periodic orbits by approximating their Poincar\'{e} mapping, whereas here we seek targets defined by macroscopic properties, and the learned solution turns out to be a recurrent solution. 

%


\subsection{Comparison to Linear Quadratic Regulator} \label{sec:LQR}

To compare our learned control policy to a conventional control method, we compare to Linear Quadratic Regulator (LQR) \citep{Hespanha2009} given the same control authority. We now consider the system state as $x$ and the control signal as $u$ where previously we referred to them as $u$ and $a$, respectively, to maintain nomenclature consistency with LQR conventions. The LQR method seeks to find a gain matrix, $\textbf{K}$, for a linear state-feedback controller, $u=-\textbf{K}x$, that minimizes the quadratic cost function $J$, 
\begin{equation}
  J = \int_{0}^{\infty} (x^T\textbf{Q}x + u^T\textbf{R}u)dt.
 \label{eq:LQRcost}
\end{equation}
for a system whose dynamics are approximated by a set of linear (or linearized) ODEs $\dot{x}=\textbf{A}x+\textbf{B}u$, where $x$ is the state of the system and $u$ the control input. We take $\textbf{Q}$, the state cost, and $\textbf{R}$, the input cost, to be the identity. Importantly, the target state of LQR i.e.~the state about which the dynamical model is linearized must be chosen a priori. Although the KSE possesses nonlinear dynamics, LQR might in certain situations be able to control the dynamics toward its target, given sufficiently close initial conditions. 



\begin{figure}[t]
 	\centering
	\begin{subfigure}[t]{0.9\textwidth}
		\includegraphics[width=\textwidth]{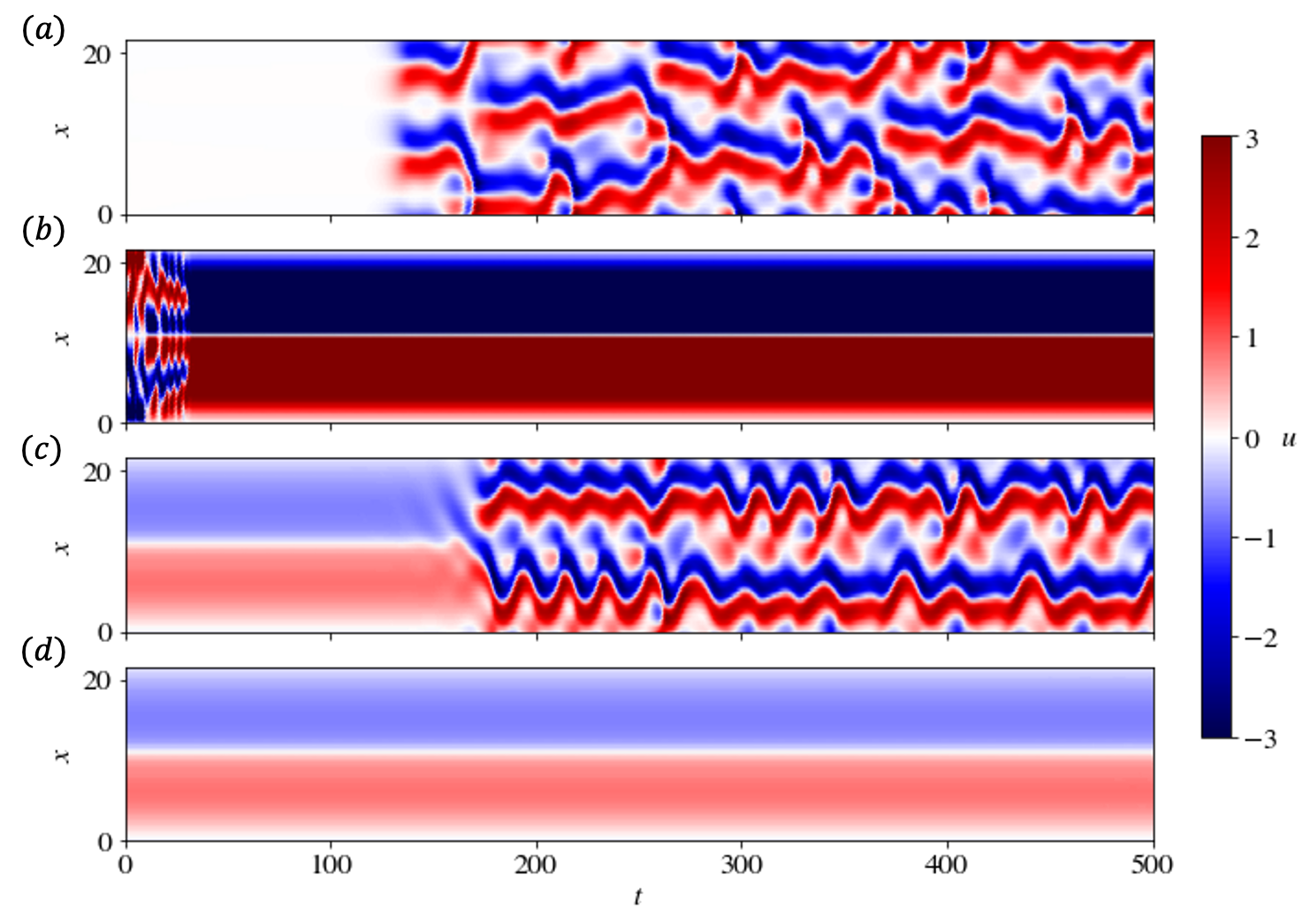}
		\phantomcaption
		\label{fig:LQRcontrol-a}
	\end{subfigure}
	\begin{subfigure}[t]{0.0\textwidth}
		\includegraphics[width=\textwidth]{figures/LQRtrajectory}
		\phantomcaption
		\label{fig:LQRcontrol-b}
	\end{subfigure}
	\begin{subfigure}[t]{0.0\textwidth}
		\includegraphics[width=\textwidth]{figures/LQRtrajectory}
		\phantomcaption
		\label{fig:LQRcontrol-c}
	\end{subfigure}
	\begin{subfigure}[t]{0.0\textwidth}
		\includegraphics[width=\textwidth]{figures/LQRtrajectory}
		\phantomcaption
		\label{fig:LQRcontrol-d}
	\end{subfigure}

	\caption[]{A trajectory initialized on the trivial solution with an infinitesimal perturbation with (a)  no additional control, $f_{LQR}=0$, and (b) with a LQR controller, $f_{LQR}$, based on the zero solution. A trajectory initialized on $E_{\alpha_{22}}$ with constant forcing $f=\alpha_{22}$ and an infinitesimal perturbation  with (c) no additional control, $f_{LQR}=0$, and (d) with a LQR controller, $f_{LQR}$, based on $E_{\alpha_{22}}$.}
	\label{fig:LQRcontrol-full}
\end{figure}

We first consider applying an LQR controller to the trivial zero solution, which in the interest of minimizing dissipation and power-input cost, is the natural target as it possesses zero dissipation and zero power input cost. The trivial zero solution is known to be linearly unstable. Shown in Fig. \ref{fig:LQRcontrol-a} is a trajectory of the trivial zero solution given an infinitesimal perturbation; it evolves to the chaotic attractor.
In this case, we find that LQR cannot stabilize the zero solution. Given the linearized KSE dynamics and the available control authority, the LQR approach fails the Popov-Belevitch-Hautus (PBH) controllability test \citep{Hespanha2009}, 


\begin{equation}
  \text{rank}[\textbf{A}-\lambda \textbf{I}\quad \textbf{B}]=n,\quad \forall \lambda\in \mathbb{C},
 \label{eq:PBHcontrollability}
\end{equation}
where $n$ is the number of rows of $\textbf{A}$. 
This indicates that LQR is not capable of transferring every state to the origin in finite time. Furthermore, it also fails the PBH stabilizability test \citep{Hespanha2009}, 
\begin{equation}
  \text{rank}[\textbf{A}-\lambda \textbf{I}\quad \textbf{B}]=\text{n},\quad \forall \lambda\in \mathbb{C}:\text{Re}[\lambda]\geq 0,
 \label{eq:PBHstabilizability}
\end{equation}
which indicates LQR is not capable of reaching the origin even given infinite time. These results suggest that  the zero solution cannot be controlled by linear means with the current actuation scheme. We comment that this inability to control the zero solution is further linked to work performed by \citep{Grigoriev1998,Grigoriev2000}, which demonstrated that periodically arranged actuator or sensor sites in systems with translational or reflection invariance can detrimentally impact controllability.

Shown in \ref{fig:LQRcontrol-b} is a trajectory initialized on the zero solution plus a small random perturbation, and controlled by the LQR controller designed based on the zero solution. We observe that this controller is unable to stabilize the trivial solution. We further note the immediate deviation from the zero solution, which is a product of a runaway controller, and emphasize that the resulting dynamics are a product of the LQR controller reaching the actuator saturation limit (which we set to be 10 times that available to the deep RL agent). We comment that although the initial short-time trajectory is asymmetric, the resulting symmetry of the trajectory is the product of the saturated controller actuating symmetrically.

 We hypothesize that a RL policy attempting to stabilize the zero solution would likely meet the same issue as the LQR approach. This may explain why the agent chooses to target the solution $E_{\alpha_{22}}$. To further investigate, we linearize the KSE, forced with $f=\alpha_{22}$ around its steady state $E_{\alpha_{22}}$, and use LQR to find the gain matrix for this system:
\begin{equation}
  a_{LQR}=-\textbf{K}(u-E_{\alpha_{22}}).
 \label{eq:LQRfeedbacksignal}
\end{equation}
Here $a_{LQR}$ are the control signals produced by LQR, which are applied to the KSE via Eq.~\ref{eq:forcingterm} as $f_{LQR}$.
This linear feedback controller is then applied to the full nonlinear KSE under constant forcing $f=\alpha_{22}$,
\begin{equation}
  u_t = -uu_{x}-u_{xx}-u_{xxxx}+f_{\alpha_{22}}+f_{LQR}.
 \label{eq:LQRaroundEalpha22}
\end{equation}
Shown in Fig. \ref{fig:LQRcontrol-c} is a trajectory initialized on $E_{\alpha_{22}}$ with constant forcing $f=\alpha_{22}$ and given an infinitesimal perturbation. In the absence of additional control, the linearly unstable solution $E_{\alpha_{22}}$ eventually evolves to the dynamics of the forced attractor. Shown in Fig. \ref{fig:LQRcontrol-d} is a trajectory initialized on $E_{\alpha_{22}}$ under constant forcing $f=\alpha_{22}$ with the LQR controller $f_{LQR}$ applied. With the addition of the LQR controller, the perturbed $E_{\alpha_{22}}$ solution can be maintained. These observations indicate that one reason why the agent learns to discover and stabilize $E_{\alpha_{22}}$ rather than the zero solution is because in the limit of approaching an equilibrium solution the agent will behave approximately linearly, and in that limit $E_{\alpha_{22}}$ can be controlled by linear means while the zero solution cannot. We reiterate that, while we have found that an LQR approach can stabilize $E_{\alpha_{22}}$, thus reducing energy consumption,  that approach needed to know the existence and structure of the steady state a priori, while the RL approach did not.

\subsection{Robustness} \label{sec:Robustness}

\begin{figure}[t]
	\begin{center}
		\includegraphics[width=0.9\textwidth]{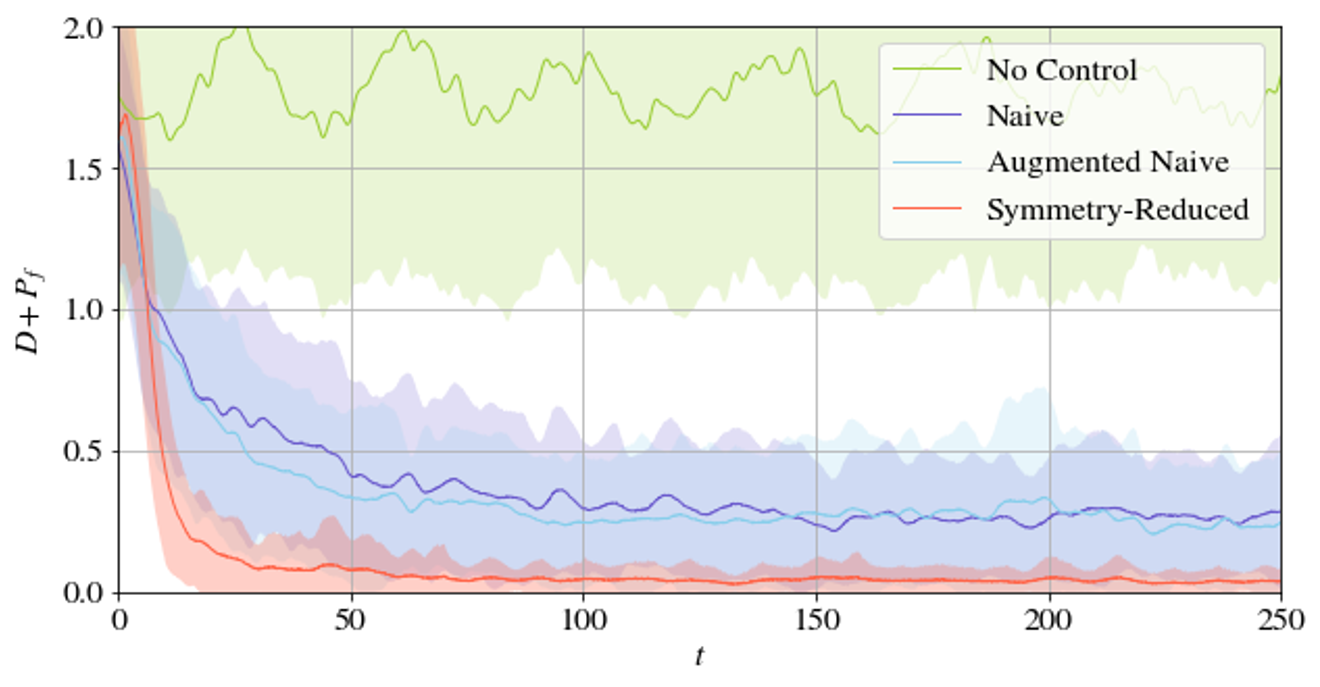}
		\caption[]{Ensemble $D+P_f$ vs. Time for 100 initial conditions controlled by: Ensemble mean $D+P_f$ of 100 trajectories controlled by: No control (green), Naive (purple), augmented naive (blue), and translation+reflection-reduced (red). All agents experience Gaussian measurement noise and actuation noise of mean zero and standard deviation $0.1$.}
		\label{fig:NoiseRobustness}
	\end{center}
\end{figure}

Chaotic systems are characterized by their sensitivity to noise and changes to system parameters. To assess the robustness of the agents to measurement and actuation noise, we tested the performance of our various RL policies, without additional training, on 100 trajectories of 250 time units with Gaussian measurement and actuation noise with zero mean and standard deviation $0.1$. The ensemble mean and standard deviation of the  $D+P_f$ trajectories of each agent type is shown in Fig. \ref{fig:NoiseRobustness}. Comparison with Fig.~\ref{fig:EnsembleDissipation} reveals that all agent have comparable performance with or without noise. In particular the symmetry-reduced agent is capable of maintaining its high performance compared to the naive and augmented naive agents.

For the KSE, altering the domain size parameter, $L$, can lead to very different dynamics. These distinct dynamics are shown in Fig. \ref{fig:BoxsizeRobustness-a} and Fig. \ref{fig:BoxsizeRobustness-b} for domain sizes of $L=21$ and $L=23$, respectively. To assess the robustness of the learned control policy to changes in $L$, the symmetry-reduced agent trained in the domain $L=22$ is applied, without additional training, to control the KSE dynamics at $L=21$ and $23$. In these experiments we maintain the same actuator jet parameters as was available in $L=22$ training (spatial Gaussian distribution and magnitude range) while maintaining equidistant placement in the new domain sizes. Shown in Fig. \ref{fig:BoxsizeRobustness-c} and Fig. \ref{fig:BoxsizeRobustness-d} are trajectories with the controller turned out at $t=100$, in domain sizes of $L=21$ and $L=23$, respectively. We observe that the agent drives both systems to equilibrium-like states similar to that found in the original domain $L=22$. Furthermore, shown in Fig.  \ref{fig:BoxsizeRobustness-e} and Fig. \ref{fig:BoxsizeRobustness-f} are $D$ and $P_f$ of these controlled trajectories; these controlled steady states again exhibit low dissipation and low power-input. In both $L=21$ and $L=23$ the targeted states are also unstable, as once the controller is switched off the systems return to their respective typical dynamics. These experiments highlight the robustness of the control policy to new unseen dynamics as well as deviations in relative control authority, as the artificial jets are smaller relative to the $L=23$ domain than in the original $L=22$ domain.

\begin{figure}[t]
 	\centering
	\begin{subfigure}[t]{0.9\textwidth}
		\includegraphics[width=\textwidth]{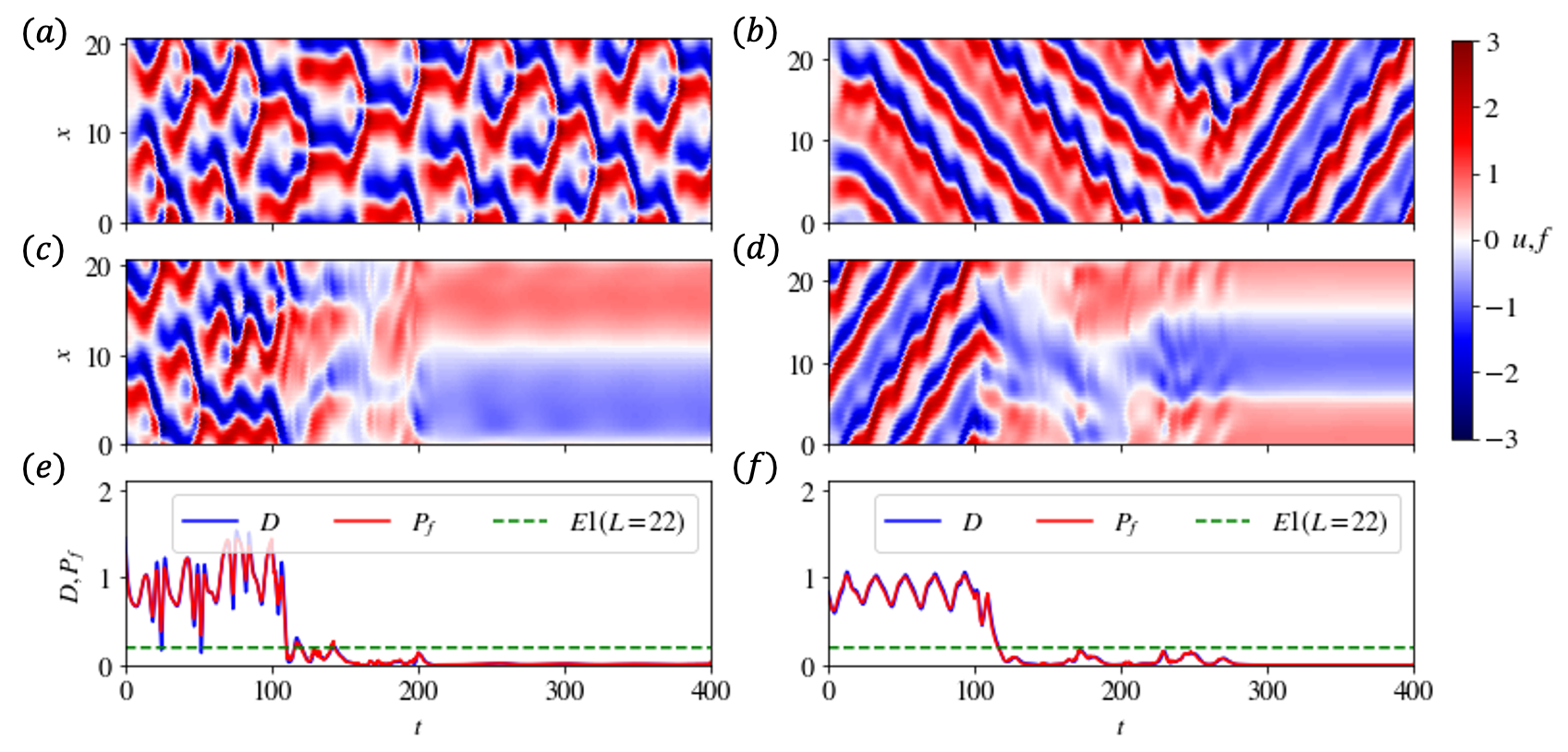}
		\phantomcaption
		\label{fig:BoxsizeRobustness-a}
	\end{subfigure}
	\begin{subfigure}[t]{0.0\textwidth}
		\includegraphics[width=\textwidth]{figures/L21L23traj}
		\phantomcaption
		\label{fig:BoxsizeRobustness-b}
	\end{subfigure}
	\begin{subfigure}[t]{0.0\textwidth}
		\includegraphics[width=\textwidth]{figures/L21L23traj}
		\phantomcaption
		\label{fig:BoxsizeRobustness-c}
	\end{subfigure}
	\begin{subfigure}[t]{0.0\textwidth}
		\includegraphics[width=\textwidth]{figures/L21L23traj}
		\phantomcaption
		\label{fig:BoxsizeRobustness-d}
	\end{subfigure}
	\begin{subfigure}[t]{0.0\textwidth}
		\includegraphics[width=\textwidth]{figures/L21L23traj}
		\phantomcaption
		\label{fig:BoxsizeRobustness-e}
	\end{subfigure}
	\begin{subfigure}[t]{0.0\textwidth}
		\includegraphics[width=\textwidth]{figures/L21L23traj}
		\phantomcaption
		\label{fig:BoxsizeRobustness-f}
	\end{subfigure}

	\caption[]{Typical dynamics for domain sizes of (a) $L=21$, (b) $L=23$. Controlled trajectory in which the $L=22$ symmetry reduced agent is applied with no additional training at $t=100$ in domain sizes of (c) $L=21$, (d) $L=23$. Dissipation and total power input as a function of time for the controlled trajectories in domain sizes of (e) $L=21$, (f) $L=23$.}
		\label{fig:BoxsizeRobustness-full}
\end{figure}


To investigate the dynamical connection between the final targeted states of the $L=21, 23$ systems and the original $L=22$ system, a two-stage continuation was performed. First, continuation in the magnitude of the forcing was performed to determine the connection between the forced systems yielding $E_{\alpha_{21}}$ and $E_{\alpha_{23}}$ and their unforced counterparts ($f=0$) of their respective domain size. The solutions found as $f$ is decreased to zero are shown in Fig. \ref{fig:ContinuationBoxSize-a} and Fig. \ref{fig:ContinuationBoxSize-b} for $L=21$ and $L=23$, respectively. These results reveal that $E_{\alpha_{21}}$ and $E_{\alpha_{23}}$ are the forced counterparts of existing equilibria in the unforced systems, which we denote as $E_{L21}$ and $E_{L23}$, respectively.

\begin{figure}[t]
 	\centering
	\begin{subfigure}[t]{0.9\textwidth}
		\includegraphics[width=\textwidth]{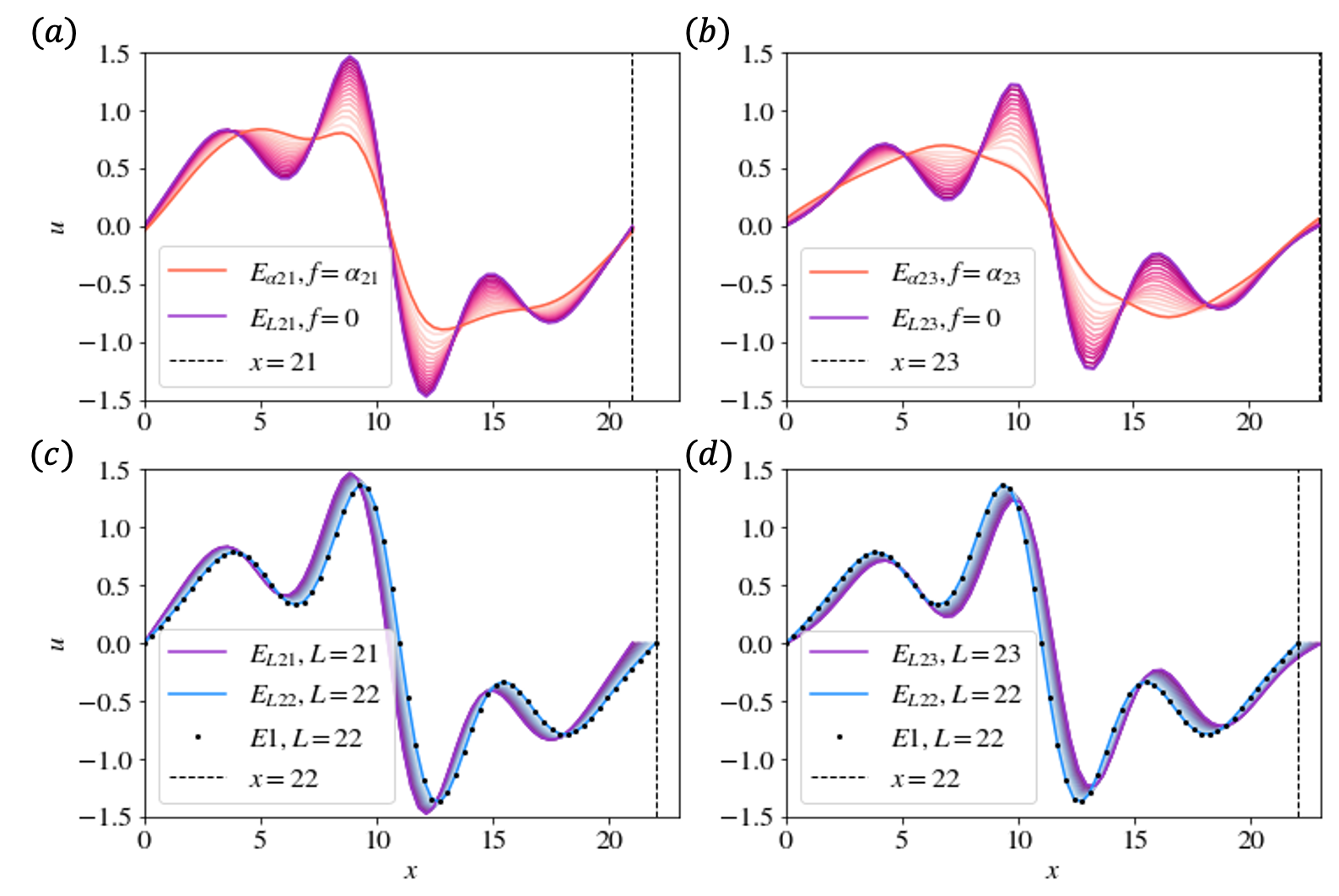}
		\phantomcaption
		\label{fig:ContinuationBoxSize-a}
	\end{subfigure}
	\begin{subfigure}[t]{0.0\textwidth}
		\includegraphics[width=\textwidth]{figures/ContInFandL}
		\phantomcaption
		\label{fig:ContinuationBoxSize-b}
	\end{subfigure}
	\begin{subfigure}[t]{0.0\textwidth}
		\includegraphics[width=\textwidth]{figures/ContInFandL}
		\phantomcaption
		\label{fig:ContinuationBoxSize-c}
	\end{subfigure}
	\begin{subfigure}[t]{0.0\textwidth}
		\includegraphics[width=\textwidth]{figures/ContInFandL}
		\phantomcaption
		\label{fig:ContinuationBoxSize-d}
	\end{subfigure}

	\caption[]{Continuation in forcing and domain size. (a) Forcing continuation from $f=\alpha_{21}$ (orange) to $f=0$ (purple) to yield $E_{\alpha_{21}}$ to $E_{L21}$. (b) Forcing continuation from $f=\alpha_{23}$ (orange) to $f=0$ (purple) to yield $E_{\alpha_{23}}$ to $E_{L23}$. (c) Domain size continuation from $L=21$ (purple) to $L=22$ (blue) to yield $E_{L21}$ to $E1$ (dots).  (d) Domain size continuation from $L=23$ (purple) to $L=22$ (blue) to yield $E_{L23}$ to $E1$ (dots).}
		\label{fig:ContinuationBoxSize-full}
\end{figure}

We next perform a second continuation, this time in the domain size, to determine the connection between $E_{L21}$, s$E_{L23}$ and the original dynamics of $L=22$. Solutions are shown in Fig. \ref{fig:ContinuationBoxSize-c} and Fig. \ref{fig:ContinuationBoxSize-d} respectively, as the domain size changes. These evolve to the $E1$ solution of $L=22$. These connections indicate that the symmetry-reduced agent is also capable of finding and stabilizing the forced $E1$ solution in domain sizes it has not seen before. Interestingly, the equilibria found by the agent, $E_{\alpha_{21}}$ and $E_{\alpha_{23}}$, are not simply spatial dilations or compressions of $E_{\alpha_{22}}$, as $E_{\alpha_{21}}$ exhibits 4 velocity peaks while $E_{\alpha_{21}}$ exhibits only two. Furthermore, we comment that the long-time mean actuation profiles utilized by the agent in the two unseen domain sizes are also distinctly different than that utilized in $L=22$, which indicates the agent is not just simply imposing the same long-time control signals it found in its original domain size of training.


\section{Conclusions} \label{Conclusions}

Although deep RL in recent years has demonstrated the capability of controlling systems with high-dimensional state-action spaces, its naive application towards spatiotemporal chaotic systems exhibiting symmetry, which encompasses many flow geometries of interest, can be limited by NN architecture and the cost of exploring the full state-action space. In this paper we proposed a modification to the general deep RL learning problem that can better learn control strategies for chaotic flow problems exhibiting symmetries by moving the learning problem into a state-action symmetry-reduced subspace.


Our method alleviates technical demands of NN architectures in existing deep RL methods such as the need for the Actor and Critic networks in DDPG to learn weight constraints that preserve equivariance and invariance, respectively. From a policy perspective, symmetry reduction alleviates the need to learn and consolidate the optimal policy for each symmetric-subspace within a single network, freeing capacity for approximating the optimal policy while maintaining a dynamically equivalent state-action mapping. As the learning problem is performed in the symmetry-reduced subspace, all training data is also generated within the symmetry-reduced subspace improving training data efficiency, as the agent no longer requires a complete exploration of the full state-action space as it would in the naive application of deep RL. Although in this work we utilized the DDPG algorithm, the commentary and conclusions drawn regarding symmetry reduction of the learning space can be extended to other deep RL methods.

We demonstrated these ideas by controlling the periodic KSE to minimize dissipation, a spatiotemporally chaotic model system for turbulence that exhibits translational and reflection symmetries. We show that by reducing the symmetry of the learning problem we can obtain faster and more consistent learning. Furthermore, we demonstrate that the control strategy found by the symmetry-reduced agent is robust to input/output noise as well system parameter perturbations. Finally, we observe that in order to achieve the objective of reduced overall power consumption, the symmetry-reduced agent discovers a low-dissipation equilibrium solution of a nontrivially forced KSE. This observation highlights a potentially important connection to effective control approaches for drag reduction in turbulent flows, as the dynamics of turbulence are organized, at least to some extent, by underlying invariant solutions known as Exact Coherent States \cite{Graham:2020ba}.
%

We further emphasize that conventional controllers typically target microscopic objectives, such as a priori known states that exhibit desirable macroscopic properties such as system dissipation, pressure drop, etc., but these a priori targets may not always be accessible with the available control authority. In our experiments here, the symmetry-reduced deep RL demonstrates the potential to serve as a discovery tool for alternative solutions with desirable macroscopic properties. These discovered states might then be utilized as alternative control targets for conventional controllers when a priori known states are inaccessible given the available control freedom.

\begin{acknowledgments}
This work was supported by AFOSR  FA9550-18-1-0174 and ONR N00014-18-1-2865 (Vannevar Bush Faculty Fellowship). We gratefully acknowledge discussions with Daniel Floryan on LQR control.
\end{acknowledgments}

\bibliography{manuscript1.bib}
\end{document}